\pdfoutput=1
\documentclass[11pt]{article}

\usepackage[]{ACL2023}

\usepackage{times}
\usepackage{latexsym}

\usepackage[T1]{fontenc}
\usepackage[utf8]{inputenc}

\usepackage{microtype}

\usepackage{inconsolata}

\usepackage{CJKutf8}
\usepackage{amsmath}
\usepackage{graphicx}
\usepackage{makecell}
\usepackage{array, multirow}
\usepackage[table, x11names, dvipsnames, xcdraw]{xcolor}
\usepackage[normalem]{ulem}
\useunder{\uline}{\ul}{}
\usepackage{booktabs}
\usepackage{nicematrix}
\usepackage{color, colortbl}
\usepackage{hhline}

\definecolor{Gray}{gray}{0.9}

\usepackage{hyperref}
\usepackage{enumitem}
\usepackage[super]{nth}
\usepackage{amsfonts}
\usepackage{subfigure}

\usepackage{pmboxdraw}

\title{ \raisebox{-0.1\height}{\includegraphics[height=0.9em]{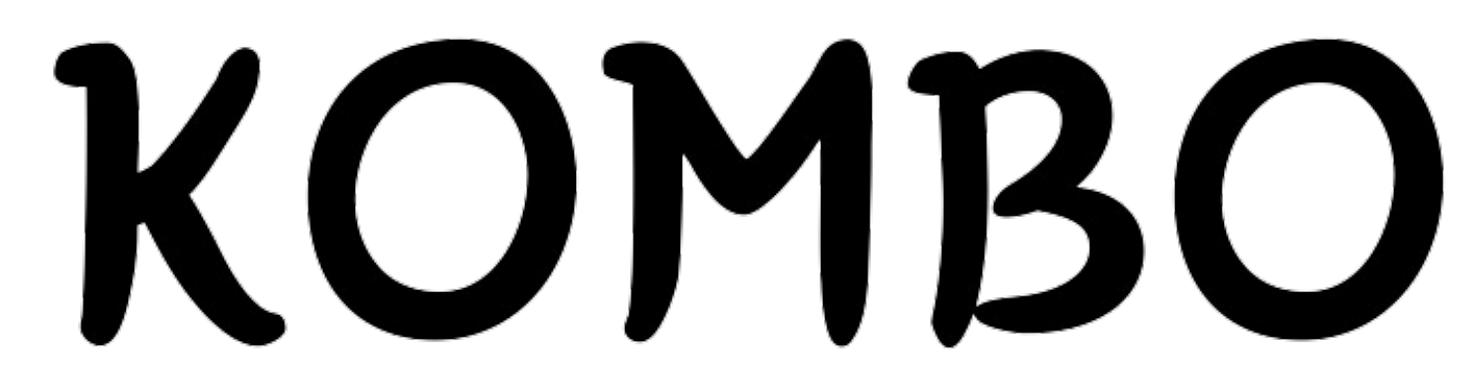}}: Korean Character Representations \\ Based on the Combination Rules of Subcharacters }

\author{
  \textbf{SungHo Kim}\textsuperscript{1}\thanks{\enspace These authors contributed equally to this work.} ,
  \textbf{Juhyeong Park}\textsuperscript{1}\footnotemark[1] ,
  \textbf{Yeachan Kim}\textsuperscript{1},
  \textbf{SangKeun Lee}\textsuperscript{1,2}
\\
  \textsuperscript{1}Department of Artificial Intelligence, Korea University, Seoul, South Korea \\
  \textsuperscript{2}Department of Computer Science and Engineering, Korea University, Seoul, South Korea
\\
  \texttt{\{sungho3268,johnida,yeachan,yalphy\}@korea.ac.kr}
}

\begin{document}

\begin{CJK}{UTF8}{mj}

\maketitle

\begin{abstract}
The Korean writing system, \textit{Hangeul}, has a unique character representation rigidly following the invention principles recorded in \textit{Hunminjeongeum}.\footnote{\textit{Hunminjeongeum} is a book published in 1446 that describes the principles of invention and usage of \textit{Hangeul}, devised by King Sejong \cite{Hunminjeongeum_Guide}.} However, existing pre-trained language models (PLMs) for Korean have overlooked these principles. In this paper, we introduce a novel framework for Korean PLMs called \raisebox{-0.1\height}{\includegraphics[height=0.9em]{assets/KOMBO.pdf}}, which firstly brings the invention principles of \textit{Hangeul} to represent character. Our proposed method, \raisebox{-0.1\height}{\includegraphics[height=0.9em]{assets/KOMBO.pdf}}, exhibits notable experimental proficiency across diverse NLP tasks. In particular, our method outperforms the state-of-the-art Korean PLM by an average of 2.11\% in five Korean natural language understanding tasks. Furthermore, extensive experiments demonstrate that our proposed method is suitable for comprehending the linguistic features of the Korean language.
Consequently, we shed light on the superiority of using subcharacters over the typical subword-based approach for Korean PLMs.
Our code is available at: \href{https://github.com/SungHo3268/KOMBO}{https://github.com/SungHo3268/KOMBO}.
\end{abstract}

\section{Introduction}
Determining the representation of a word is the first stepping stone in building pre-trained language models. The predominant approach in English has attempted to decompose each word (e.g., \textit{pureness}) into subwords (e.g., \textit{pure + ness}) based on the frequency of the words, such as byte-pair encoding (BPE) \cite{sennrich-etal-2016-neural}, WordPiece \cite{schuster2012japanese}, and SentencePiece \cite{kudo-richardson-2018-sentencepiece}. Notably, even for the Korean language, which has a significantly different linguistic structure from English, the subword-based approach has been widely adopted for Korean PLMs \cite{park-etal-2020-empirical, klue-dataset}.

However, the Korean writing system (known as \textit{Hangeul}) has a unique property in representing letters. Unlike English, which typically adheres to a "word-subword-character" structure, Korean includes an additional "subcharacter" level consisting of \textit{chosung} (initial consonants), \textit{joongsung} (vowels), and \textit{jongsung} (final consonants). This leads to a "word-subword-character-subcharacter" structure, giving rise to distinct Korean linguistic features. This structure is even problematic when generating compound words, such as 기찻길$_{\text{train track}}$.\footnote{기찻길$_{\text{train track}}$ is formed from 기차$_{\text{train}}$ and 길$_{\text{track}}$, where a subcharacter `ㅅ' is inserted between two words.} Despite these nuances in subcharacters with the same meaning, subword-based methods are blind to this information, typically parsing the compound word as separate tokens: 기, 찻, and 길. Such structural differences raise questions about the suitability of the subword-based method for the Korean PLMs, considering that it may overlook important linguistic information beyond characters.

In this work, given that it is crucial to ground the linguistic nuances and unique structures of the Korean language, we draw our attention towards subcharacter in building Korean PLMs. To accurately reflect the linguistic information of subcharacters, we bring the historical document \textit{Hunminjeongeum}, which provides comprehensive insights into the design principles of subcharacters and their combination rules.
The following are the two essential statements related to subcharacters \cite{Hunminjeongeum_Haerye}:
\begin{itemize}[leftmargin=10px]
    \item \textbf{[\textit{Design of the Letters}]} Chosung (initial consonant), joongsung (vowel), and jongsung (final consonant) are combined to form a single character representing a syllable.
    \item \textbf{[\textit{Combination of the Letters}]} Chosung can be placed above joongsung, or it can be positioned to the left of joongsung. Jongsung is placed below chosung and joongsung.
\end{itemize} 

Building on these principles, we propose a novel framework for Korean PLMs, referred to as \raisebox{-0.1\height}{\includegraphics[height=0.9em]{assets/KOMBO.pdf}} (\textbf{KO}rean character representations based on the co\textbf{MB}inati\textbf{O}n rules of subcharacters). To instill the design principles and combination rules of subcharacters into PLMs, \raisebox{-0.1\height}{\includegraphics[height=0.9em]{assets/KOMBO.pdf}} starts with subcharacters as the initial representations. These representations are progressively combined to form a character through a merging layer guided by the combination principles. Additionally, we also introduce a masking strategy tailored to subcharacters, aimed at learning the structural knowledge of characters during pre-training. Consequently, such a comprehensive and fundamental approach enables PLMs to better comprehend the unique structure and composition of the Korean language.

To demonstrate the efficacy of the proposed method, we perform extensive experiments over a wide range of NLP tasks and compare ours with a variety of Korean tokenization methods in building PLMs. The results convincingly show that considering subcharacter, especially for Jamo, in a principled manner brings substantial improvement to Korean PLMs. Moreover, the in-depth analysis supports that \raisebox{-0.1\height}{\includegraphics[height=0.9em]{assets/KOMBO.pdf}} can better understand the Korean features, such as conjugations and offensive language, compared to existing methods. In summary, the contributions of this paper include the following:
\begin{itemize}
[leftmargin=15px]
    \item We present \raisebox{-0.1\height}{\includegraphics[height=0.9em]{assets/KOMBO.pdf}}, a novel framework for Korean PLMs grounded in the invention principles of \textit{Hangeul} as specified in \textit{Hunminjeongeum}.
    
    \item We integrate the design of characters and combination rules into neural language models, marking a novel exploration in Korean PLMs.
    
    \item We demonstrate that considering the structure of \textit{Hangeul} through invention principles leads to remarkable performance on a wide range of NLP tasks, thereby shedding light on the potential of Jamo units for Korean PLMs.
\end{itemize}

\begin{figure*}[h]
    \centering
    \includegraphics[width=1.\textwidth]{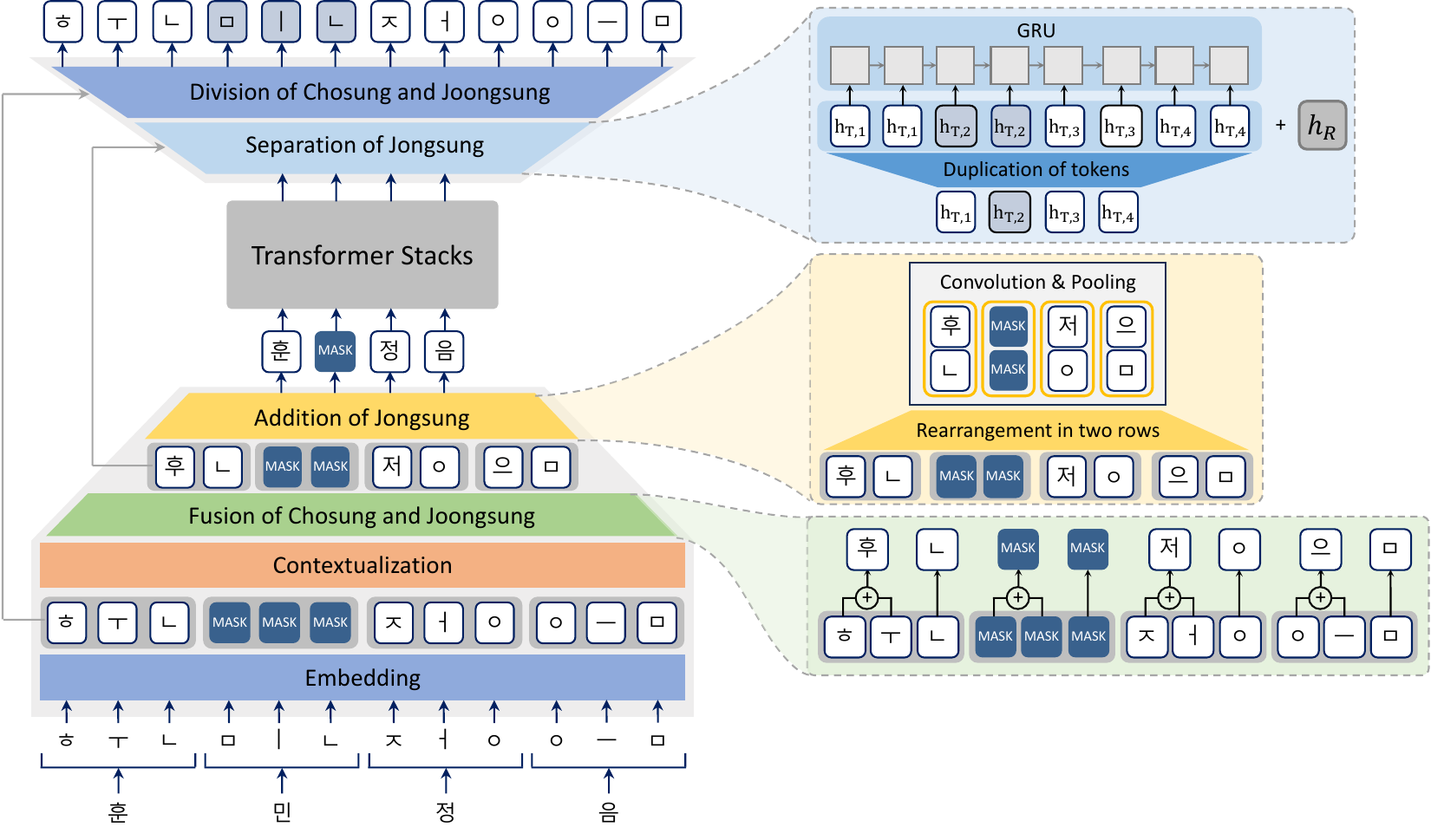}
    \caption{Overall illustration of \raisebox{-0.1\height}{\includegraphics[height=0.9em]{assets/KOMBO.pdf}} where the input is "훈민정음$_\textit{Hunminjeongeum}$" which has four characters and twelve subcharacters. The model starts with the twelve subcharacters and progressively combines them to construct four characters based on the combination principles, e.g., ㅎ+ㅜ+ㄴ $\rightarrow$ {\textbf{후}+ㄴ}$\rightarrow$ \textbf{훈}. After going through the transformer stack, the model is trained to predict the consecutive subcharacters with the restoration layers.}
    \label{figure_1}
    % \vspace{-0.3cm}
\end{figure*}

\section{Related Work}
\subsection{Korean Pre-trained Language Models}
Pre-trained language models have understood the word by splitting it into various atomic units. Similar to other languages, Korean PLMs have also adopted tokenizing sentences into character-level \cite{Cho2019Investigating}, subword-level \cite{Lee2018Comparing, park-etal-2018-subword}, or have split by whitespace \cite{9761258}. Moreover, to consider the morphologically rich feature of the Korean language, \citet{lee-shin-2021-korean} and \citet{moon-etal-2022-openkorpos} have utilized a morpheme analyzer \cite{Kudo2005MeCabY} to tokenize sentences.
\citet{park-etal-2020-empirical}, \citet{kim-etal-2021-changes}, and \citet{klue-dataset} have presented morpheme-aware subword tokenization, which is the subword-level tokenization approach applying a morpheme analyzer before implementing BPE. Within this diverse range of units, most of the current Korean PLMs have used morpheme-aware subword units as basic tokens to split the sentence because of their high performance. However, subword-level methodologies have been insufficient in addressing the distinctive linguistic attributes \cite{albright2009predicting, park-shin-2018-grapheme} inherent in the Korean language, which are influenced by the presence of subcharacters specific to \textit{Hangeul}.

\subsection{Korean Subcharacter Units}
To consider the unique structural information of Korean letters, there have been some trials to tokenize the word into subcharacters for word embeddings. \citet{stratos-2017-sub} and \citet{moon-okazaki-2020-jamo} have split words into the subcharacter unit Jamo, which is used as a basic subcharacter of \textit{Hangeul} letters. Additionally, \citet{kim-etal-2022-break} have decomposed Jamo into smaller tokens, BTS units, which are inspired by the invention principle of \textit{Hangeul}. Depending on the decomposition level, they have discerned BTS units into three different units, such as Stroke, Cji, and BTS. However, most of the previous works have been exploited only in static word embeddings, leaving a research gap in the application in PLMs, such as BERT \cite{devlin2019bert}. 
Therefore, in this work, we propose a novel framework for Korean PLMs based on the design principles and combination rules of subcharacters as mentioned in \textit{Hunminjeongeum}.

\section{\raisebox{-0.1\height}{\includegraphics[height=0.9em]{assets/KOMBO.pdf}}}\label{KOMBO}
In this section, we elaborate \raisebox{-0.1\height}{\includegraphics[height=0.9em]{assets/KOMBO.pdf}} in details for Korean PLMs. The framework begins with taking subcharacters as initial representations (\S\ref{representation}). These subcharacter representations are progressively combined to form a character based on the combination rules (\S\ref{combination}). After passing through the transformer blocks (\S\ref{transformer}), the subcharacter representations are reconstructed to perform token-level objectives (\S\ref{seperation}). To learn the structural knowledge of subcharacters during pre-training, we also introduce the span-subcharacter masking strategy (\S\ref{span character masking}). Figure~\ref{figure_1} illustrates the overall procedures.

\subsection{Initial Representations of \raisebox{-0.1\height}{\includegraphics[height=0.9em]{assets/KOMBO.pdf}}}
\label{representation}
\raisebox{-0.1\height}{\includegraphics[height=0.9em]{assets/KOMBO.pdf}} begins with treating subcharacters as atomic units for representing a word. While there are a number of ways to represent subcharacters (e.g., Stroke, Cji, BTS), we primarily use \textbf{Jamo} in the following sections for the sake of simplicity in explanation.\footnote{Details regarding BTS units are provided in Appendix~\ref{sec:appendixA}.}

\paragraph{Subcharacter Tokenization} Each input character is decomposed into chosung, joongsung, and jongsung. Chosung and joongsung are essential components to create a character, whereas jongsung is an optional component. To represent the absence of jongsung, we use the special empty token ({\tiny \color{gray} \pmboxdrawuni{2583}}), ensuring that each character is always represented with three Jamo units (e.g., a character `차$_\text{car}$' is decomposed to ㅊ,ㅏ, and {\tiny \color{gray} \pmboxdrawuni{2583}}). We denote the subcharacter input sequence as $\mathbf{x} = \{x_1, x_2, \cdots, x_N\}\in\mathbb{R}^N$, where $N$ is the number of subcharacters.

\paragraph{Subcharacter Embedding} We project the input sequence $\mathbf{x}$ onto the embedding space. The projected token representations are denoted as follows:
\begin{align} \label{eq:1}
    \mathbf{e} = \text{Embedding$(\mathbf{x})$}\in \mathbb{R}^{N \times D}
\end{align}
where $D$ is the dimension of embeddings.

\subsection{Subcharacter Combination}
\label{combination}
\paragraph{Contextualization} Building on the principles of combination, we integrate subcharacter representations to construct a character. However, simply merging these representations without considering context and sequence fails to capture the intricate composition of characters. Therefore, we apply contextualization to the subcharacters using shallow local transformer blocks, subsequently followed by Gated Recurrent Units (GRU) \cite{cho-etal-2014-learning}. Through the contextualization layers, the subcharacter representations are derived as follows:
\begin{equation} \label{eq:2}
    \mathbf{h} = \text{GRU}(\text{LocalTransformer}(\mathbf{e}))\in\mathbb{R}^{N \times D}
\end{equation}

\paragraph{Merging Jamo units}
\label{merging}
In the following sections for the intuitive illustration, we refer to the structural representation $h_i\in\mathbf{h}$ as chosung $h_{\text{I}, k}$, joongsung $h_{\text{V}, k}$, and jongsung $h_{\text{F}, k}$, respectively:
\begin{equation} \label{eq:3}
    h_i=
    \begin{cases}
    h_{\text{I}, k} & \mbox{if } i = 3k-2  \\
    h_{\text{V}, k} & \mbox{if } i = 3k-1  \\
    h_{\text{F}, k} & \mbox{if } i = 3k  
    \end{cases}
\end{equation} where each $k \in [1, N/3]$ corresponds to a character index in the sequence.

Subsequently, to form a character representation, we combine denoted subcharacters in two steps. Depending on the order of combining subcharacters as referred to \textit{\textbf{Design of the Letters}}, we first merge chosung and joongsung through element-wise summation of $\textbf{h}_\text{I}=\{h_{\text{I},k}\}$ and $\textbf{h}_\text{V}=\{h_{\text{V},k}\}$. 
\begin{align} 
\mathbf{h}_\text{I+V}&=\mathbf{h}_\text{I} + \mathbf{h}_\text{V} \;\; \in\mathbb{R}^{{N\over{3}}\times D} \label{eq:4}
    % \mathbf{h}_\text{F}&=\{h_{\text{F},k}\} \quad\quad\quad\quad\;\, \in\mathbb{R}^{{N\over{3}}\times D}
\end{align}
The second merging step is to combine the resultant representations with jongsung.
Considering that jongsung is always positioned below chosung and joongsung, as mentioned in \textit{\textbf{Combination of the Letters}}, we treat it by performing vertical concatenation of jongsung representations $\textbf{h}_\text{F}=\{h_{\text{F},k}\}$ with the combined chosung and joongsung representations $\mathbf{h}_\text{I+V}$. Formally, this can be represented as follows:
\begin{equation} \label{eq:5}
    \textbf{h}_\text{R}=
    \begin{bmatrix}
        \mathbf{h}_\text{I+V} \\
        \mathbf{h}_\text{F}
    \end{bmatrix}
    \in\mathbb{R}^{2\times{N\over{3}}\times D}
\end{equation}

To generate the final subcharacter jongsung based on the merged representations, we perform convolution and pooling operations\footnote{We heuristically explore and use $(2\times 1)$ kernels with the stride of one to perform convolution.} over the concatenated representations $\textbf{h}_\text{R}$.
\begin{equation} \label{eq:6}
    \mathbf{h}_\text{C} = \text{AvgPool}(\text{Conv}(\mathbf{h}_\text{R}))\in\mathbb{R}^{{N\over{3}}\times D}
\end{equation}
This results in a dense character representation grounded in the combination rules of subcharacters.

\subsection{Transformer Stack}
\label{transformer}
On top of the merged representations, we employ a series of transformer, consisting of $L$ layers, to achieve contextualization as follows:
\begin{equation} \label{eq:7}
    \mathbf{h}'_\text{C} = \text{Transformer}_{L} (\mathbf{h}_\text{C})
\end{equation}
The configuration of these transformer layers follows the same design as if the BERT model \cite{devlin2019bert}. Notably, it is crucial to highlight that our framework is primarily designed to manipulate token representations, allowing for seamless integration with various transformer-based architectures, extending beyond BERT.

\subsection{Subcharacter Restoration}
\label{seperation}
In sentence-level tasks, the first token (i.e., [CLS]) from the transformer stack is utilized to perform the specified task. However, token-level classification necessitates a fine-grained sequential output that aligns with the vocabulary, requiring subcharacter representations. We thus introduce the restoration layers after the transformer stack, which convert the character representations back into the constituent subcharacters. 

The reconstruction proceeds by reversing the process of the subcharacter combination. First, the character representations ($\mathbf{h}'_\text{C}$ or $\mathbf{h}'_\text{R}$) is duplicated by the number of tokens used in each subcharacter combination process (Eq.~\eqref{eq:6} or Eq.~\eqref{eq:4}). Inspired by \textit{U-Net} \cite{ronneberger2015u}, the duplicated representations are subsequently combined with the original subcharacter representations (i.e., $\mathbf{h}_\text{R}$ or $\mathbf{h}$) for better restoration. Finally, by leveraging the GRU layer, we ensure the continuity between subcharacters during the reconstruction process. In summary, the restoration process\footnote{Comparison with the variation of restoration process is provided in Appendix~\ref{sec:appendixB.1}.} can be formulated as follows:
\begin{align}
    \mathbf{h}'_\text{R}&=\text{GRU}(\text{Repeat}(\mathbf{h}'_\text{C}) + \mathbf{h}_\text{R}) \;\,
    \in \mathbb{R}^{{2\over{3}}N\times D} \label{eq:8} \\
    \mathbf{h}'&=\text{GRU}(\text{Repeat}(\mathbf{h}'_\text{R}) + \mathbf{h}) \quad 
    \in \mathbb{R}^{N \times D} \label{eq:9}
\end{align}

\subsection{Span-Subcharacter Masking}\label{span character masking}

Furthermore, to learn the linguistic structure within characters and their corresponding subcharacters, we also introduce a span-subcharacter masking strategy inspired by SpanBERT \cite{joshi-etal-2020-spanbert}. Specifically, instead of masking tokens at arbitrary positions, we mask out the consecutive three subcharacters (corresponding to chosung, joongsung, and jongsung) of each character for the objective of MLM. Such a masking strategy encourages the model to learn the compositional relationships between subcharacters in the pre-training phase, thereby leading to subcharacter-aware pre-trained language models.

\begin{table*}[h]
\centering
\setlength{\tabcolsep}{3pt}        % 열 여백
\renewcommand{\arraystretch}{1.}    % 행 여백
\begin{tabular}{clccccccccccc}
\toprule
\multirow{2.5}{*}{Model} & \multirow{2.5}{*}{Tokenization} & \multirow{2.5}{*}{\begin{tabular}[l]{@{}c@{}} Vocab\\ Size\end{tabular}} & KorQuAD & \multicolumn{2}{c}{KorNLI} & \multicolumn{2}{c}{KorSTS} & \multicolumn{2}{c}{NSMC} & \multicolumn{2}{c}{PAWS-X} \\ 
\cmidrule(lr){4-4} \cmidrule(lr){5-6} \cmidrule(lr){7-8} \cmidrule(lr){9-10} \cmidrule(lr){11-12}
 & & & Dev(EM/ F1) & Dev & Test & Dev & Test & Dev & Test & Dev & Test \\ 
\toprule
\multirow{9}{*}{$\text{BERT}$} & Stroke & 130 & 38.40/ 50.42 & 57.78 & 57.66 & 67.99 & 67.62 & 87.02 & 86.84 & 56.49 & 54.67 \\ 
 & Cji & 136 & 32.64/ 44.48 & 56.00 & 57.37 & 63.86 & 64.38 & 86.85 & 86.70 & 55.77 & 54.67 \\
 & BTS & 112 & 18.78/ 30.30 & 50.77 & 50.52 & 56.63 & 60.14 & 86.14 & 86.23 & 54.45 & 55.33 \\
 & Jamo & 170 & 55.73/ 68.90 & 62.27 & 64.73 & 77.22 & 72.96 & 87.84 & 87.78 & 60.40 & 59.09 \\
 \cmidrule(){2-12}
 & Character & 2K & 55.67/ 74.67 & 72.51 & 72.59 & 83.98 & 76.34 & {\ul 89.03} & 88.89 & 69.51 & 68.70 \\
 & Morpheme & 32K & 65.28/ 80.73 & 73.90 & 72.57 & 82.19 & 74.35 & 87.55 & 87.40 & {\ul 72.97} & 66.66 \\
 & Subword & 32K & {\ul 70.50}/ {\ul 84.23} & 73.14 & 73.32 & 83.80 & 76.41 & {\ul 89.03} & {\ul 88.91} & 72.31 & {\ul 68.88} \\
 & MorSubword & 32K & 66.20/ 80.69 & {\ul 73.91} & {\ul 73.76} & {\ul 84.26} & {\ul 77.29} & \textbf{89.59} & \textbf{89.40} & 71.23 & 68.14 \\
 & Word & 64K & 2.45/ 8.86 & 66.77 & 65.45 & 72.20 & 65.50 & 74.91 & 74.16 & 65.42 & 61.16 \\ 
\midrule
\multirow{4}{*}{\raisebox{-0.1\height}{\includegraphics[height=0.9em]{assets/KOMBO.pdf}}} & Stroke & 130 & 64.64/ 75.59 & 70.31 & 70.58 & 82.42 & 75.55 & 87.65 & 87.47 & 65.25 & 63.79 \\
 & Cji & 136 & 68.84/ 79.68 & 71.74 & 72.13 & 83.37 & 75.28 & 86.55 & 88.08 & 66.56 & 66.11 \\
 & BTS & 112 & 58.24/ 69.49 & 66.77 & 65.97 & 79.73 & 72.90 & 86.68 & 86.40 & 62.86 & 62.49 \\
 & Jamo & 170 & \textbf{77.47}/ \textbf{86.30} & \textbf{74.22} & \textbf{73.77} & \textbf{84.47} & \textbf{77.43} & 88.71 & 88.70 & \textbf{73.66} & \textbf{70.88} \\
\bottomrule
\end{tabular}
\caption{Performance of various tokenization methods for PLMs on standard Korean datasets. The evaluation metrics for each task are as follows: KorQuAD: Exact Match/ macro F1, KorNLI: accuracy (\%), KorSTS:  100 $\times$ Spearman correlation, NSMC: accuracy (\%), PAWS-X: accuracy (\%). The best and second-best results are highlighted in \textbf{boldface} and \underline{underline}, respectively.}
\vspace{-0.3cm}
\label{table_1}
\end{table*}

\section{Experiments}
In this section, we conduct a comparative analysis between the proposed method and existing Korean tokenization models on both standard Korean datasets and noisy Korean datasets.

\subsection{Experimental Setup}
\paragraph{Models} We compare our proposed methods with Korean PLMs using 9 distinct tokenization methods, including BTS units (Stroke, Cji, and BTS), Jamo, Character, Morpheme, Subword, MorSubword (short for Morpheme-aware Subword~\cite{park-etal-2020-empirical, klue-dataset, kim-etal-2021-changes}), and Word units. We uniformly apply BERT to all PLMs, using the same configuration as BERT$_\textit{base}$ with 12 transformer blocks. Depending on the types of subcharacters, we categorize our methods as \raisebox{-0.1\height}{\includegraphics[height=0.9em]{assets/KOMBO.pdf}}$_\text{Stroke}$, \raisebox{-0.1\height}{\includegraphics[height=0.9em]{assets/KOMBO.pdf}}$_\text{Cji}$, \raisebox{-0.1\height}{\includegraphics[height=0.9em]{assets/KOMBO.pdf}}$_\text{BTS}$, and \raisebox{-0.1\height}{\includegraphics[height=0.9em]{assets/KOMBO.pdf}}$_\text{Jamo}$. We match the size of all our models with the state-of-the-art Korean PLM, MorSubword, by adjusting the number of local transformer blocks in the contextualization layer.

\paragraph{Pre-training}~We pre-trained all models for 1M steps on Masked Language Modeling (MLM) and Next Sentence Prediction (NSP) tasks as BERT. We used the Korean Wiki corpus and the Namuwiki corpus, 6.2 GB, including about 46M sentences in total. The details of data preprocessing and hyperparameter settings are explained in Appendix~\ref{sec:appendixC}.

\begin{table*}[h]
\centering
\setlength{\tabcolsep}{5pt}        % 열 여백
\renewcommand{\arraystretch}{1.}    % 행 여백
\begin{tabular}{clccccccccccc}
\toprule
\multirow{2.5}{*}{Model} & \multirow{2.5}{*}{Tokenization} & \multirow{2.5}{*}{Clean} & \multicolumn{2}{c}{Insertion} & \multicolumn{2}{c}{Transposition} & \multicolumn{2}{c}{Substitution} & \multicolumn{2}{c}{Deletion} \\ 
 \cmidrule(lr){4-5} \cmidrule(lr){6-7} \cmidrule(lr){8-9} \cmidrule(lr){10-11}
 & & & +20\% & +40\% & +20\% & +40\% & +20\% & +40\% & +20\% & +40\% \\ 
\toprule
\multirow{4}{*}{$\text{BERT}$} & Jamo & 64.73 & 62.87 & 60.54 & 56.38 & 51.31 & 62.60 & 61.36 & 56.47 & 51.29 \\ 
 & Character & 72.59 & \underline{69.41} & \underline{66.91} & 61.59 & \underline{54.79} & \underline{68.36} & \underline{64.57} & 62.10 & 54.71 \\
 & Subword & 73.32 & 67.22 & 63.43 & 62.45 & \textbf{54.83} & 67.17 & 61.83 & 62.14 & \textbf{55.46} \\
 & MorSubword & \underline{73.76} & 68.60 & 64.63 & \underline{62.77} & 54.36 & 67.62 & 62.07 & \underline{62.57} & 54.86 \\
\midrule
\raisebox{-0.1\height}{\includegraphics[height=0.9em]{assets/KOMBO.pdf}} & Jamo & \textbf{73.77} & \textbf{70.63} & \textbf{67.73} & \textbf{62.82} & 54.54 & \textbf{70.83} & \textbf{68.06} & \textbf{63.21} & \underline{54.94} \\

\bottomrule
\end{tabular}
\caption{Performance on KorNLI with typo. We measure the sensitivity to typo rate using four different typo generation methods. The best and second-best results are highlighted in \textbf{boldface} and \underline{underline}, respectively.}
\vspace{-0.3cm}
\label{table_2}
\end{table*}

\paragraph{Evaluation}~ Pre-trained models were individually fine-tuned on each downstream task dataset. We report all experimental results as the average values obtained from three random seeds. To verify the effectiveness of the proposed method in standard Korean dataset, we evaluated the models on five Korean NLU tasks, including machine reading comprehension (KorQuAD 1.0), natural language inference (KorNLI), semantic textual similarity (KorSTS), sentiment analysis (NSMC), and paraphrase identification (PAWS-X). We provide detailed explanations about the overall data information and hyperparameter settings in Appendix~\ref{sec:appendixD.1}.

Moreover, to assess the robustness of models in noisy Korean settings, we synthetically injected the typos into the Korean NLU datasets, such as KorNLI, KorSTS, NSMC, and PAWS-X. We randomly generated typos using four different typo methods, following \citet{zhuang-zuccon-2021-dealing}; Insertion: randomly add a letter which is adjacent of the letter on the keyboard, Transposition: randomly switches a letter with one of its neighbor letter, Substitution: randomly changes a letter with one of its neighbor letters on the keyboard, and Deletion: drops a random letter. The other settings, unless specified, follow the same manner as Korean NLU tasks.

\subsection{Experiment Results}
\subsubsection{Standard Korean Datasets}
\label{KoreanNLU}
In Table~\ref{table_1}, our proposed method \raisebox{-0.1\height}{\includegraphics[height=0.9em]{assets/KOMBO.pdf}} consistently outperforms BERT models, which use the same subcharacter tokenization as the initial representation, for all tasks with marginal increases, averaging more than 9\%. We observe that, among ours, \raisebox{-0.1\height}{\includegraphics[height=0.9em]{assets/KOMBO.pdf}}$_\text{Jamo}$ presents the most overwhelming performances, while \raisebox{-0.1\height}{\includegraphics[height=0.9em]{assets/KOMBO.pdf}} for BTS units shows lagging performances. We suspect that this is due to the truncated initial inputs at subcharacter level by a setting of maximum sequence length, giving rise to short context. \raisebox{-0.1\height}{\includegraphics[height=0.9em]{assets/KOMBO.pdf}}$_\text{Jamo}$ outperforms the state-of-the-art model MorSubword across all tasks except for NSMC. Despite using only 0.53\% of the size of the static embeddings compared to the MorSubword model, our method achieves higher performance by an average of 2.11\%. These admirable results provide valuable insight into the prospective potential of Jamo, taking over the trend of current subword-based Korean PLMs.

\subsubsection{Noisy Korean Datasets}
\label{NoisyKorean}
\begin{figure}[t]
    \centering
    \includegraphics[width=0.50\textwidth]{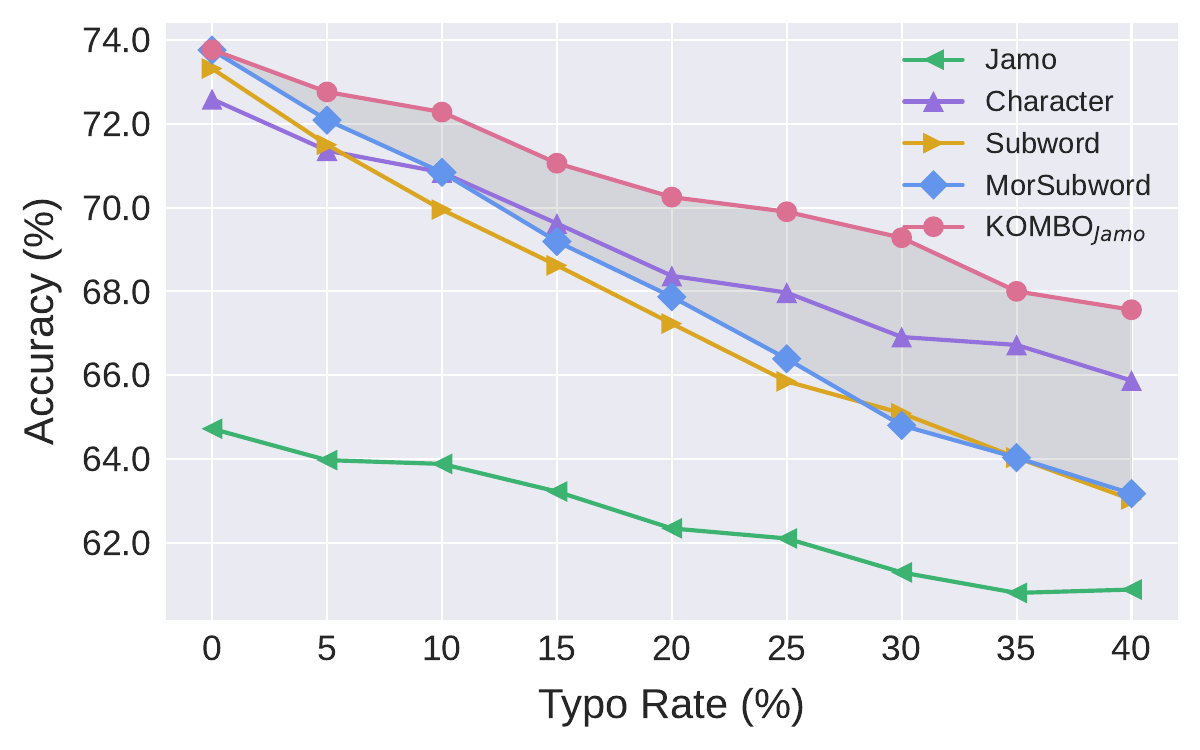}
    \vspace{-0.5cm}
    \caption{Evaluation results on KorNLI with random typo methods by increasing the typo rate. For better visualization of gap, we fill the area between our proposed method and the state-of-the-art baseline in gray.}
    \vspace{-0.3cm}
    \label{figure_2}
\end{figure}

We evaluate the robustness of models to typos in two experimental settings. The first is the random typo setting, where we raise the typos by randomly selecting typo methods among Insertion, Transposition, Substitution, and Deletion, starting from the rate of 0\% and increasing it by 5\% up to 40\%. In Figure~\ref{figure_2}, the baselines using larger token units, such as Subword and MorSubword, are highly vulnerable to typo, whereas the models using smaller token units, such as Jamo and Character, generally exhibit strong robustness with a gradual decline. Moreover, aligning with the trends shown in the baselines, our proposed model \raisebox{-0.1\height}{\includegraphics[height=0.9em]{assets/KOMBO.pdf}}$_\text{Jamo}$ also exhibited strong robustness to typo. Furthermore, we can observe that the performance gap between \raisebox{-0.1\height}{\includegraphics[height=0.9em]{assets/KOMBO.pdf}}$_\text{Jamo}$ and the state-of-the-art model is getting wider as if the typo rate increases, again demonstrating the strong robustness of our proposed method.
The second setting is that we inject only one type of typo error among four different typo methods. In Table~\ref{table_2}, we observe that our method is more powerful in the substitution typo method, which is the most similar setting to real-world typographical errors \cite{jeon2010analyzing}. We report more results for other tasks in Appendix~\ref{sec:appendixD.2}.

\begin{table*}[h]
\centering
\setlength{\tabcolsep}{5pt}         % 열 여백
\renewcommand{\arraystretch}{1.}    % 행 여백
\begin{tabular}{lccccccc}
\toprule
\multirow{2.5}{*}{Condition} & \multicolumn{2}{c}{KorNLI} & \multicolumn{2}{c}{KorSTS} & \multicolumn{2}{c}{PAWS-X} & \multicolumn{1}{c}{Total} \\ 
\cmidrule(lr){2-3} \cmidrule(lr){4-5} \cmidrule(lr){6-7} \cmidrule(lr){8-8}
 & \multicolumn{1}{c}{Dev} & \multicolumn{1}{c}{Test} & \multicolumn{1}{c}{Dev} & \multicolumn{1}{c}{Test} & \multicolumn{1}{c}{Dev} & \multicolumn{1}{c}{Test} & \multicolumn{1}{c}{Avg} \\
\toprule
\raisebox{-0.1\height}{\includegraphics[height=0.9em]{assets/KOMBO.pdf}}$_\text{Jamo}$ & \textbf{74.22} & 73.77 & \textbf{84.47} & \textbf{77.43} & \textbf{73.66} & \textbf{70.88} & \textbf{75.74}\\
\midrule
\multicolumn{8}{c}{Subcharacter Combination}\\
\midrule
w/o Contextualization & 52.70 & 54.28 & 73.11 & 67.57 & 54.76 & 54.77 & 59.53 \\
w/o Merging Subcharacters & 71.45 & 71.68 & 82.54 & \underline{76.27} & 71.97 & 68.42 & 73.72 \\
w/o Addition of Jongsung & 73.40 & 73.69 & \underline{84.17} & 75.94 & \underline{73.17} & \underline{70.72} & \underline{75.18} \\
w/o Span-Subcharacter Masking & 72.29 & 73.08 & 83.58 & 76.13 & 72.07 & 70.21 & 74.56 \\
\midrule
\multicolumn{8}{c}{Kernel Size in Addition of Jongsung}\\
\midrule
w/ (2x2) kernel & 73.70 & \textbf{75.38} & 83.04 & 75.40 & 65.07 & 63.13 & 72.62 \\
w/ (2x3) kernel & 73.31 & 74.68 & 82.62 & 75.88 & 62.73 & 61.34 & 71.76 \\
w/ (2x1)+(2x2) kernel & \underline{73.78} & \underline{74.84} & 82.95 & 75.27 & 64.46 & 62.17 & 72.25 \\
\bottomrule
\end{tabular}
\caption{Ablation results for various components of \raisebox{-0.1\height}{\includegraphics[height=0.9em]{assets/KOMBO.pdf}}$_\text{Jamo}$.
The first row is the best setting of \raisebox{-0.1\height}{\includegraphics[height=0.9em]{assets/KOMBO.pdf}}$_\text{Jamo}$, which merges all subcharacters sequentially.
The best and second-best results are highlighted in \textbf{boldface} and \underline{underline}, respectively.}
\label{table_3}
\end{table*}

\begin{table*}[h]
\centering
\setlength{\tabcolsep}{3.5pt}         % 열 여백
\renewcommand{\arraystretch}{1.1}    % 행 여백
\begin{tabular}{lccccccc}
\toprule
\multirow{2.5}{*}{Method} & \multicolumn{2}{c}{KorNLI} & \multicolumn{2}{c}{KorSTS} & \multicolumn{2}{c}{PAWS-X} & \multicolumn{1}{c}{Total} \\ 
\cmidrule(lr){2-3} \cmidrule(lr){4-5} \cmidrule(lr){6-7} \cmidrule(lr){8-8}
 & \multicolumn{1}{c}{Dev} & \multicolumn{1}{c}{Test} & \multicolumn{1}{c}{Dev} & \multicolumn{1}{c}{Test} & \multicolumn{1}{c}{Dev} & \multicolumn{1}{c}{Test} & \multicolumn{1}{c}{Avg} \\
\toprule
\raisebox{-0.1\height}{\includegraphics[height=0.9em]{assets/KOMBO.pdf}}$_\text{Jamo}$ & \textbf{74.22} & \textbf{73.77} & \textbf{84.47} & \textbf{77.43} & \textbf{73.66} & \textbf{70.88} & \textbf{75.74}\\
\midrule
Attention Pooling* \cite{NEURIPS2020_2cd2915e} & 63.91 & \underline{63.38} & 80.58 & \underline{72.43} & 54.93 & 55.37 & 65.10 \\
Linear Pooling* \cite{nawrot-etal-2022-hierarchical} & \underline{63.93} & 62.59 & \underline{80.65} & 72.02 & \underline{55.52} & \underline{55.97} & \underline{65.11} \\
\bottomrule
\end{tabular}
\caption{Comparison between \raisebox{-0.1\height}{\includegraphics[height=0.9em]{assets/KOMBO.pdf}}$_\text{Jamo}$ and other downsampling methods used in English (denoted as *). Note that we apply each downsampling methods on  \raisebox{-0.1\height}{\includegraphics[height=0.9em]{assets/KOMBO.pdf}}$_\text{Jamo}$ instead of proposed subcharacter combination methods for comparison. The best and second-best results are highlighted in \textbf{boldface} and \underline{underline}, respectively.}
\vspace{-0.3cm}
\label{table_4}
\end{table*}

\section{Ablations}
\label{Ablations}
In Table~\ref{table_3}, we present ablation experiments on the subcharacter combination methods of \raisebox{-0.1\height}{\includegraphics[height=0.9em]{assets/KOMBO.pdf}}. The subsequent segments present the experimental results corresponding to the modifications in the structure of subcharacter combination and kernel size used in merging jongsung, respectively. In Table~\ref{table_4}, to investigate the effect of Korean character structural knowledge, we compare our subcharacter combination method with other downsampling methods generally used in English.

\subsection{Impact of Contextualization}
\label{Contextualization}
We evaluate our proposed model by removing the contextualization layer before merging embeddings. The omission of the contextualization layer has resulted with a significant drop in performance, which clearly indicates the assisting role of contextualization in character representations. Through these results, we insist that complementing small embeddings is necessary to enhance the performance of subcharacter models.

\subsection{Impact of Span-Subcharacter Masking}
\label{masking_ablation}
Instead of applying the span-subcharacter masking strategy for MLM, we adopt the token-level, i.e., subcharacter masking strategy. We observe that using the span-subcharacter masking strategy enhances the quality of the combined character representation while increasing the average performance by 1\%. Through the results, we verify the efficacy of span-subcharacter masking to learn the structural information of characters.

\subsection{Impact of Merging Subcharacters}
\label{merginig_subcharacter}
The model that merges subcharacters shows a significant performance improvement, averaging approximately 2.02\% higher than the model without subcharacter merging. Furthermore, removing the addition of jongsung results in a performance decrease, averaging around 0.56\%. These consistently lower performances underscore the importance of the subcharacter merging steps.

\subsection{Impact of Kernel Size}
\label{kernel_size}
By changing the size and number of the kernels, we explore the most effective kernel size in our subcharacter combination method. The experimental results show that using a (2x1) size kernel, which looks at each character individually, yields the best performance across all tasks except the KorNLI test set. This indicates that the best understanding of Korean is obtained when focusing on the character level, which is the composite unit of Hangeul.

\begin{figure*}[h]
    \centering
    \includegraphics[width=1.\textwidth]{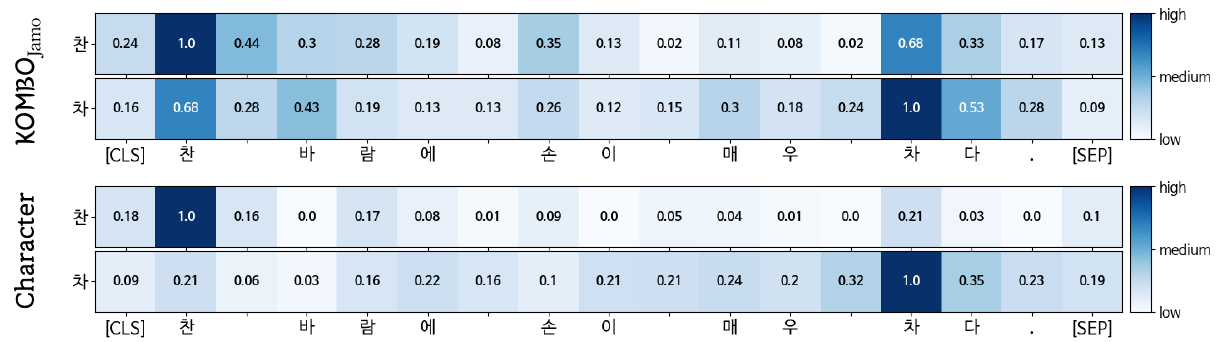}
    \caption{Visualization of the character representations.
    Given the target sentence as "\textbf{찬} 바람에 손이 매우 \textbf{차}다. (My hands are very \textbf{cold} in the \textbf{cold} wind.)", the histogram represents the cosine similarities between each of the two embeddings of the character `찬$_\text{cold}$' and `차$_\text{cold}$' and all characters in the target sentence.}
    \label{figure_3}
\end{figure*}

\subsection{Impact of Korean Character Structure}
\label{FusingMethod}
In Table~\ref{table_4}, to explore the effect of recognition of character structure in Korean, we compare our method with other English downsampling methods \cite{NEURIPS2020_2cd2915e, nawrot-etal-2022-hierarchical}, which do not consider the unique Korean character structure. Instead of our subcharacter combination method, we opt to employ each downsampling method to combine subcharacter embeddings. Both methods show significant performance drops by an average of over 10\% compared to our method. We demonstrate that considering the structure of \textit{Hangeul} is a keystone to comprehending the Korean language.

\begin{table*}[h]
\centering
\setlength{\tabcolsep}{5pt}        % 열 여백
\renewcommand{\arraystretch}{1.}    % 행 여백
\begin{tabular}{clccccccccc}
\toprule
\multirow{2.5}{*}{Model} & \multicolumn{1}{l}{\multirow{2.5}{*}{Tokenization}}  & \multicolumn{3}{c}{BEEP!} & \multicolumn{3}{c}{K-MHaS} & \multicolumn{3}{c}{KOLD} \\
\cmidrule(lr){3-5} \cmidrule(lr){6-8} \cmidrule(lr){9-11}
 & & P & R & F1 & P & R & F1 & P & R & F1 \\
\toprule
\multirow{4}{*}{$\text{BERT}$}
 & Jamo         & 87.05 & 69.78 & 77.43 & 77.85 & 73.27 & 75.13 & 39.03 & 47.96 & 41.95 \\
 & Character    & \underline{89.62} & 71.27 & 79.40 & \underline{78.58} & 75.24 & 76.45 & \underline{46.97} & 52.82 & \underline{49.12} \\
 & Subword      & 87.30 & 71.49 & 78.61 & 77.01 & 75.28 & 75.90 & 45.88 & \underline{53.59} & 48.66 \\
 & MorSubword   & 89.09 & \underline{72.45} & \underline{79.91} & 77.99 & \underline{75.63} & \underline{76.58} & 46.70 & 52.45 & 48.91 \\
\midrule
$\text{\raisebox{-0.1\height}{\includegraphics[height=0.9em]{assets/KOMBO.pdf}}}$ & Jamo & \textbf{90.66} & \textbf{74.28} & \textbf{81.57} & \textbf{79.21} & \textbf{76.15} & \textbf{77.48} & \textbf{47.22} & \textbf{55.79} & \textbf{50.42} \\
\bottomrule
\end{tabular}
\caption{Evaluation results for the robustness of the models on three Korean offensive language datasets. 
We evaluate the model reporting performance metrics, including Precision (P), Recall (R),
and macro F1. The best and second-best results are highlighted in \textbf{boldface} and \underline{underline}, respectively.}
\vspace{-0.3cm}
\label{table_5}
\end{table*}

\section{Analysis}
\label{Analysis}
In this section, we analyze two Korean linguistic characteristics. The primary objective is to investigate the resilience of predicates against variations arising from Korean conjugations, while the secondary aim involves evaluating the model's robustness to Korean offensive languages.

\subsection{Robustness to Character Conjugations}
\label{Conjugation}
Korean, as an agglutinative language, presents a multitude of conjugational variations occurring at both the character and subcharacter levels through inserting, deletion, or substitution of endings, resulting in alterations in parts-of-speech (POS) or tense. For instance, the verb `차다$_\text{ be cold}$', composed of the stem `차$_\text{cold}$' and the ending  `다$_\text{ be}$', can transform into the adjective `찬$_\text{cold}$' by substituting the ending `다$_\text{ be}$' with the subcharacter `ㄴ'. Consequently, a profound understanding of the subcharacter structure within \textit{Hangeul} is essential for proficiently handling Korean conjugations.
We compare the vanilla character model with \raisebox{-0.1\height}{\includegraphics[height=0.9em]{assets/KOMBO.pdf}}$_\text{Jamo}$, as illustrated in Figure~\ref{figure_3}. For the given target sentence, we extract embeddings for the characters `찬' and `차' from both models. Subsequently, we compute cosine similarities between these two embeddings and all character embeddings in the target sentence, visualizing the results on a heatmap. The vanilla character model with static embeddings distinguishes `찬$_\text{cold}$' and `차$_\text{cold}$' as distinct entities, whereas \raisebox{-0.1\height}{\includegraphics[height=0.9em]{assets/KOMBO.pdf}}$_\text{Jamo}$ recognizes them as semantically similar characters, displaying significantly higher similarity scores for both embeddings (0.68) compared to the character model's score of 0.21. Further instances are provided in Appendix~\ref{sec:appendixE}.

\subsection{Robustness to Korean Offensive Language}
\label{offensive_language}
% \vspace{-0.3cm}
\textit{\textbf{Warning}: This section contains several offensive statements.} \vspace{0.2cm} \\
To evaluate the robustness of our proposed method to offensive language, we experiment on the Korean offensive language datasets BEEP! \cite{moon-etal-2020-beep}, K-MHaS \cite{lee-etal-2022-kmhas}, and KOLD \cite{jeong-etal-2022-kold}.
Details for data and experimental settings are presented in Appendix~\ref{sec:appendixD.3}.
In Table~\ref{table_5}, \raisebox{-0.1\height}{\includegraphics[height=0.9em]{assets/KOMBO.pdf}}$_\text{Jamo}$ shows the highest macro F1 score across all tasks, demonstrating its robustness on Korean offensive wordings. Interestingly, we find that the subword-based models struggle with the Korean offensive wordings; for example, they tokenize the compound word `개새끼$_\text{puppy}$', which is combined `개$_\text{dog}$' and `새끼$_\text{pup}$', into independent three characters `개', `새', and `끼'.~\footnote{In Korean, `개새끼$_\text{puppy}$' is used as an insult to refer to someone in a disrespectful manner.}

\section{Conclusion}
\label{Conclusion}
Although there is crucial linguistic information and clear invention principles in Korean letters, the existing Korean PLMs have overlooked them and just opted to use subword-level tokenization methods due to their high performance.
In this paper, we first bring attention to the overlooked design principles and combination rules of subcharacters, as specified in \textit{Hunminjeongeum}.
We have introduced a novel framework for Korean PLMs called \raisebox{-0.1\height}{\includegraphics[height=0.9em]{assets/KOMBO.pdf}}, which generates character representations following the design and combination rules of subcharacters.
We have demonstrated the efficacy of \raisebox{-0.1\height}{\includegraphics[height=0.9em]{assets/KOMBO.pdf}} on both standard Korean datasets and noisy Korean datasets by outperforming various Korean tokenization baselines on most of the tasks we evaluated. Additionally, through diverse ablations and analyses, we have shown that our proposed method improves the quality of character representations and has a more robust adaptability to Korean conjugations and offensive language. These convincing outcomes of \raisebox{-0.1\height}{\includegraphics[height=0.9em]{assets/KOMBO.pdf}}$_\text{Jamo}$ following the invention principles of \textit{Hangeul} can thereby provide an inspiring insight for the prospective potential of Jamo unit applying in Korean PLMs.

\section*{Limitations}
While our proposed methodology demonstrates its suitability for Korean natural language processing, it comes with some limitations for future research.
Although our proposed \raisebox{-0.1\height}{\includegraphics[height=0.9em]{assets/KOMBO.pdf}} is designed with an encoder transformer model, our framework is primarily designed to manipulate token representations, allowing for seamless integration with various transformer-based architectures, extending beyond Encoder Models. Therefore, we believe that its dynamic ability to generate appropriate character embeddings for each input also works well in generative modeling, too. We leave the exploration of its value in generative models for future research.

In this work, we limit our focus to Korean representations up to the character-level, as our main objective is to incorporate Korean subcharacter units into PLMs following the principles of \textit{Hangeul}. However, given the morphologically intricate nature of the Korean language, considering representations up to the subword-level is also imperative. With our hierarchical combination approach demonstrating superior performance at the character level, we anticipate that it may serve as a precursor for future methodologies transitioning from character-level to subword-level representations in Korean language processing.

\section*{Ethics Statement}
We only use three previously collected or synthetically generated Korean offensive language benchmarks, which are annotated with humans who have verified their qualifications. We strictly follow the data usage agreements for each public dataset we implement in this paper. We also mention a warning statement in \S\ref{offensive_language}, where the offensive statements are directly used in this paper. Additionally, we recognize the potential that the high performance in Korean offensive language also means the model might be biased toward misused or offensive language. However, we believe that the high robustness of our method is aligned with the advantages due to the deep comprehension of Korean character structure, not the familiarity with toxic wording.

\section*{Acknowledgements}
This work was supported by the Basic Research Program through the National Research Foundation of Korea (NRF) grant funded by the Korea government (MSIT) (2021R1A2C3010430) and Institute of Information \& Communications Technology Planning \& Evaluation (IITP) grant funded by the Korea government (MSIT) (No.RS-2019-II190079, Artificial Intelligence Graduate School Program (Korea University)).

\bibliography{custom}
\bibliographystyle{acl_natbib}

\clearpage

\appendix

\section*{Appendix}

\section{\raisebox{-0.1\height}{\includegraphics[height=0.9em]{assets/KOMBO.pdf}} for BTS units}
\label{sec:appendixA}
\begin{figure*}[h]
    \centering
    \includegraphics[width=1.\textwidth]{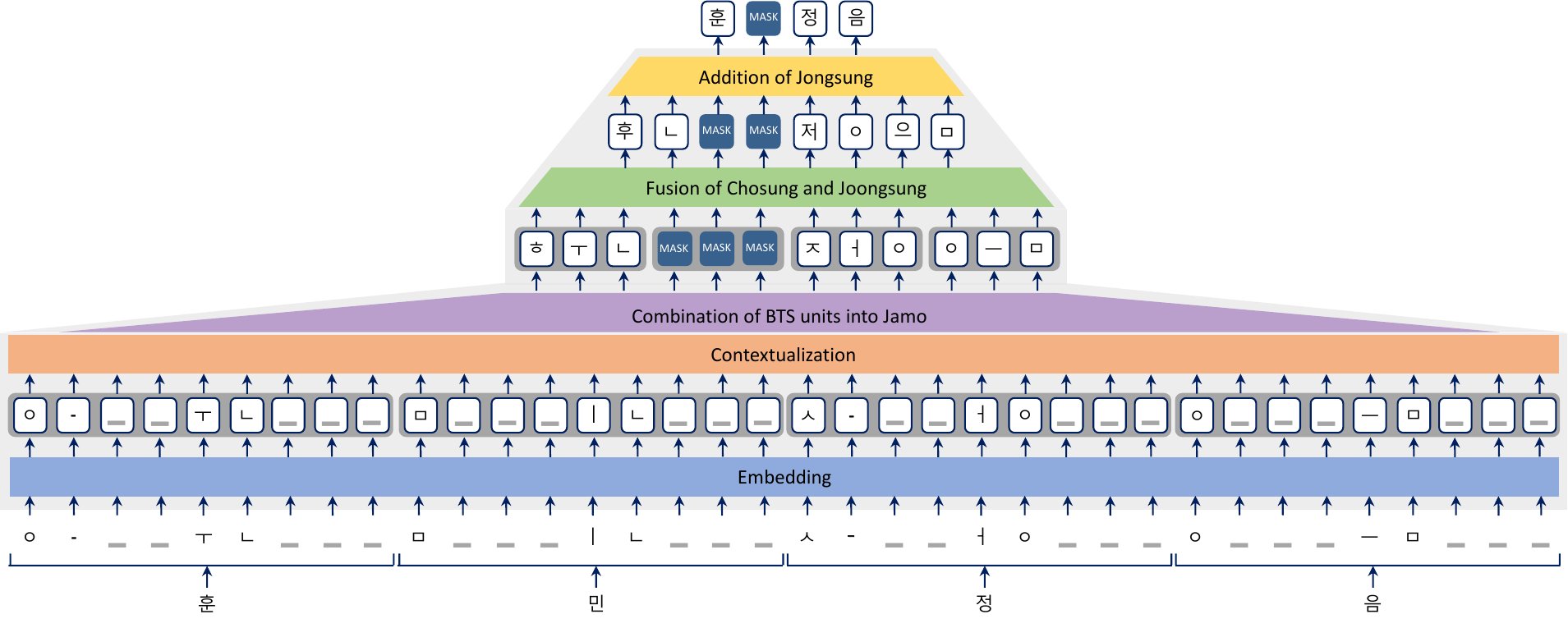}
    \caption{Detailed description of the subcharacter combination method of \raisebox{-0.1\height}{\includegraphics[height=0.9em]{assets/KOMBO.pdf}}$_\text{Stroke}$. There is one more merging layer to merge BTS units to Jamo, unlike \raisebox{-0.1\height}{\includegraphics[height=0.9em]{assets/KOMBO.pdf}}$_\text{Jamo}$.}
    \label{figure_4}
\end{figure*}

\citet{kim-etal-2022-break} introduced Basic, Tiniest Subword (BTS) units for the Korean language, inspired by the invention principle of \textit{Hangeul}. BTS units comprise 5 basic consonants and 3 basic vowels, defined as atomic units in \textit{Hunminjeongeum}. The decomposition of Korean characters into BTS units is referred to as BTS decomposition, where the consonant is split into a maximum of 4 subcharacters (e.g., a consonant `ㅋ' is decomposed into `ㄱ' and `-'), and the vowel is split into a maximum of 5 subcharacters (e.g., a vowel `ㅏ' is decomposed into `l' and `$\cdot$'). Additionally, there are two variant decomposition units: the consonant-only BTS decomposition (denoted as Stroke) and the vowel-only BTS decomposition (denoted as Cji, short for Cheonjiin).
In this paper, we consider all applications for BTS units as subcharacter units and present as \raisebox{-0.1\height}{\includegraphics[height=0.9em]{assets/KOMBO.pdf}}$_\text{Stroke}$, \raisebox{-0.1\height}{\includegraphics[height=0.9em]{assets/KOMBO.pdf}}$_\text{Cji}$, and \raisebox{-0.1\height}{\includegraphics[height=0.9em]{assets/KOMBO.pdf}}$_\text{BTS}$. To simplify, we collectively call these three models as \raisebox{-0.1\height}{\includegraphics[height=0.9em]{assets/KOMBO.pdf}}$_\text{BTS units}$.
We illustrate \raisebox{-0.1\height}{\includegraphics[height=0.9em]{assets/KOMBO.pdf}}$_\text{Jamo}$ in Figure~\ref{figure_4}, as a representative example among \raisebox{-0.1\height}{\includegraphics[height=0.9em]{assets/KOMBO.pdf}}$_\text{BTS units}$.

\paragraph{Subcharacter Tokenization}~Unlike Jamo units, which have three subcharacters for each character, BTS units decompose Jamo into smaller subcharacters.
Depending on the type of BTS unit, the number of tokens comprising one character is different. Stroke has 9, Cji has 7, and BTS has 13 tokens for each character. To ensure that the number of tokens forming each Jamo is always maximum, we use the special empty token ({\tiny \color{gray} \pmboxdrawuni{2583}}).

\paragraph{Subcharacter Embedding}~
By the same process in Eq.~\eqref{eq:1}, we project the subcharacter sequence onto the embedding space, denoted as:
\begin{equation} \label{eq:10}
    \textbf{e}=\{e_1, e_2, \cdots, e_N\} \in \mathbb{R}^{N\times D}
\end{equation} where $N$ is the number of subcharacters, and $D$ is the dimension of embeddings.

\paragraph{Contextualization}~To capture the intricate composition of characters, we apply contextualization to the subcharacters, same as in Eq.~\eqref{eq:2}.
\begin{equation} \label{eq:11}
    \mathbf{h} = \text{GRU}(\text{LocalTransformer}(\mathbf{e}))\in\mathbb{R}^{N \times D}
\end{equation}

\paragraph{Merging Jamo units}~The contextualized subcharacter representations are combined to form a character representation similar to \S\ref{merging}. Before merging chosung, joongsung, and jongsung, there is a preceding step combining the subcharacter representations into the representations of chosung, joongsung, and jongsung. We primarily use Stroke (consonant-only BTS decomposition) in the following sections for the sake of simplicity in explanation. For each $k\in [1, N/9]$, $h_{\text{I},k}, h_{\text{V},k}$, and $h_{\text{F},k}$ in Eq.~\eqref{eq:3} can be denoted as follows:

\begin{align}
    h_{\text{I}, k} =& \sum_{j=1}^{4}h_{9(k-1) + j} \label{eq:12} \\
    h_{\text{V}, k} =& \sum_{j=5}^{5}h_{9(k-1) + j} \label{eq:13} \\
    h_{\text{F}, k} =& \sum_{j=6}^{9}h_{9(k-1) + j} \label{eq:14}
\end{align}
Then, we sum the combined chosung and joongsung representations to form the intermediate representation $\mathbf{h}_\text{I+V}$ with the same process as in Eq.~\eqref{eq:4}. 
\begin{equation}\label{eq:15}
    \mathbf{h}_\text{I+V}=\mathbf{h}_\text{I}+\mathbf{h}_\text{V}
   \;\; \in\mathbb{R}^{{N\over{9}}\times D}
\end{equation}
To consider the position of jongsung, below chosung and joongsung, we vertically concatenate the combined representations $\mathbf{h}_\text{I+V}$ and jongsung $\mathbf{h}_\text{F}$, represented in Eq.~\eqref{eq:5}.
\begin{equation}\label{eq:16}
    \mathbf{h}_\text{R}=
    \begin{bmatrix}
        \mathbf{h}_\text{I+V} \\
        \mathbf{h}_\text{F}
    \end{bmatrix}
    \in\mathbb{R}^{2\times{{N\over{9}}}\times D} 
\end{equation}
The resultant representations are then performed using the operations in Eq.~\eqref{eq:6} to generate the character representation.
\begin{equation}\label{eq:17}
    \mathbf{h}_\text{C}= \text{AvgPool}(\text{Conv}(\mathbf{h}_\text{R}))
    \in\mathbb{R}^{{N\over{9}}\times D} 
\end{equation}

\paragraph{Subcharacter Restoration}~ We also reconstruct the outputs of the transformer stack into subcharacters. Due to the simplicity of the addition of subcharacters into Jamo, we directly decompose intermediate representations into BTS units, jumping the decomposition step from Jamo to BTS units.
Thus, the process of subcharacter restoration is almost the same as in Eq.~\eqref{eq:8} and Eq.~\eqref{eq:9}, but different in the size of the vectors. More formally,

\begin{align}
    \mathbf{h}'_\text{R}&=\text{GRU}(\text{Repeat}(\mathbf{h}'_\text{C}) + \mathbf{h}_\text{R})
    \;\, \in \mathbb{R}^{{2N\over{9}}\times D} \label{eq:18} \\
    \mathbf{h}'&=\text{GRU}(\text{Repeat}(\mathbf{h}'_\text{R}) + \mathbf{h}) 
    \quad \in \mathbb{R}^{N\times D} \label{eq:19}
    % \vspace{-1cm}
\end{align}

\begin{table*}[h]
\setlength{\tabcolsep}{5pt}        % 열 여백
\renewcommand{\arraystretch}{1.}    % 행 여백
\center
% \resizebox{\columnwidth}{!}{%
\begin{tabular}{lccccccc}
\toprule
\multirow{2.5}{*}{Restoration} & \multicolumn{2}{c}{KorNLI} & \multicolumn{2}{c}{KorSTS} & \multicolumn{2}{c}{PAWS-X} & \multicolumn{1}{c}{Total} \\
\cmidrule(lr){2-3} \cmidrule(lr){4-5} \cmidrule(lr){6-7} \cmidrule(lr){8-8}
             & Dev     & Test     & Dev     & Test     & Dev     & Test   & Avg    \\ 
\toprule
Linear       & 73.09$_{\pm{0.77}}$ & 73.96$_{\pm{0.14}}$ & \underline{84.42}$_{\pm{0.30}}$ & 76.23$_{\pm{0.45}}$ & 71.51$_{\pm{0.78}}$ & \textbf{71.30}$_{\pm{0.22}}$ & 75.09 \\
Linear+HR    & 73.69$_{\pm{0.13}}$ & \textbf{74.32}$_{\pm{0.25}}$ & 83.80$_{\pm{0.38}}$ & 76.43$_{\pm{0.62}}$ & 72.80$_{\pm{1.00}}$ & 70.54$_{\pm{0.32}}$ & \underline{75.26} \\
Linear+HR+RC & 73.71$_{\pm{0.08}}$ & \underline{74.22}$_{\pm{0.24}}$ & 84.15$_{\pm{0.15}}$ & \underline{77.25}$_{\pm{0.53}}$ & 72.62$_{\pm{0.69}}$ & 69.08$_{\pm{0.43}}$ & 75.17 \\
GRU          & \underline{74.04}$_{\pm{0.34}}$ & 74.20$_{\pm{0.22}}$ & 83.91$_{\pm{0.23}}$ & 75.91$_{\pm{1.19}}$ & 72.24$_{\pm{1.28}}$ & 68.63$_{\pm{2.00}}$ & 74.82 \\
GRU+HR       & 73.66$_{\pm{0.70}}$ & 73.88$_{\pm{0.28}}$ & 84.24$_{\pm{0.18}}$ & 76.40$_{\pm{0.42}}$ & \underline{73.04}$_{\pm{0.94}}$ & 70.35$_{\pm{0.60}}$ & \underline{75.26} \\
\cellcolor{Gray}GRU+HR+RC    &\cellcolor{Gray}\textbf{74.22}$_{\pm{0.45}}$ &\cellcolor{Gray}73.77$_{\pm{0.08}}$ & \cellcolor{Gray}\textbf{84.47}$_{\pm{0.31}}$ & \cellcolor{Gray}\textbf{77.43}$_{\pm{0.15}}$ & \cellcolor{Gray}\textbf{73.66}$_{\pm{0.69}}$ & \cellcolor{Gray}\underline{70.88}$_{\pm{0.58}}$ & \cellcolor{Gray}\textbf{75.74} \\
\bottomrule
\end{tabular}
\caption{Comparison of the variation of restoration process on KorNLI, KorSTS, and PAWS-X tasks. HR means restoring the subcharacter embedding hierarchically, and RC means residual connection. The best and second-best results are highlighted in \textbf{boldface} and \underline{underline}, respectively. Gray area indicates the best Decomposing method.}
\label{table_6}
\end{table*}

\section{Additional Ablations}

\subsection{Impact of Subcharacter Restoration}
\label{sec:appendixB.1}
We explore the effects of the various subcharacter restoration methods. Table~\ref{table_6} shows the experimental results. We observe that leveraging only a single GRU layer does not help subcharacter reconstruction. However, the GRU layers in hierarchical manners are effective for increasing the granularity of character-level sequences. Moreover, the residual connection, which combines duplicated representations with the original subcharacter embeddings, significantly enhances the performance when combined with the GRU layer, whereas it does not provide any advantages in the case of the linear layer.

\section{Details of Pre-training}
\label{sec:appendixC}
\paragraph{Data Preprocessing}~ We use the Korean Wiki corpus and the Namuwiki corpus, extracted from dump data using data extractors such as WikiExtractor~\footnote{\href{https://github.com/attardi/wikiextractor}{WikiExtractor}} and NamuWikiExtractor~\footnote{\href{https://github.com/jonghwanhyeon/namu-wiki-extractor/}{NamuWikiExtractor}}. After applying the data extractor to dump data, we clean the data by removing irregular empty spaces and parsing traces, such as HTML tags. We only retain Korean and English text and punctuation with regular expression. As a result, we can get a total of 6.2 GB, including about 46 million sentences.

\begin{figure}[h]
    \centering
    \subfigure[]{\includegraphics[width=0.45\textwidth]{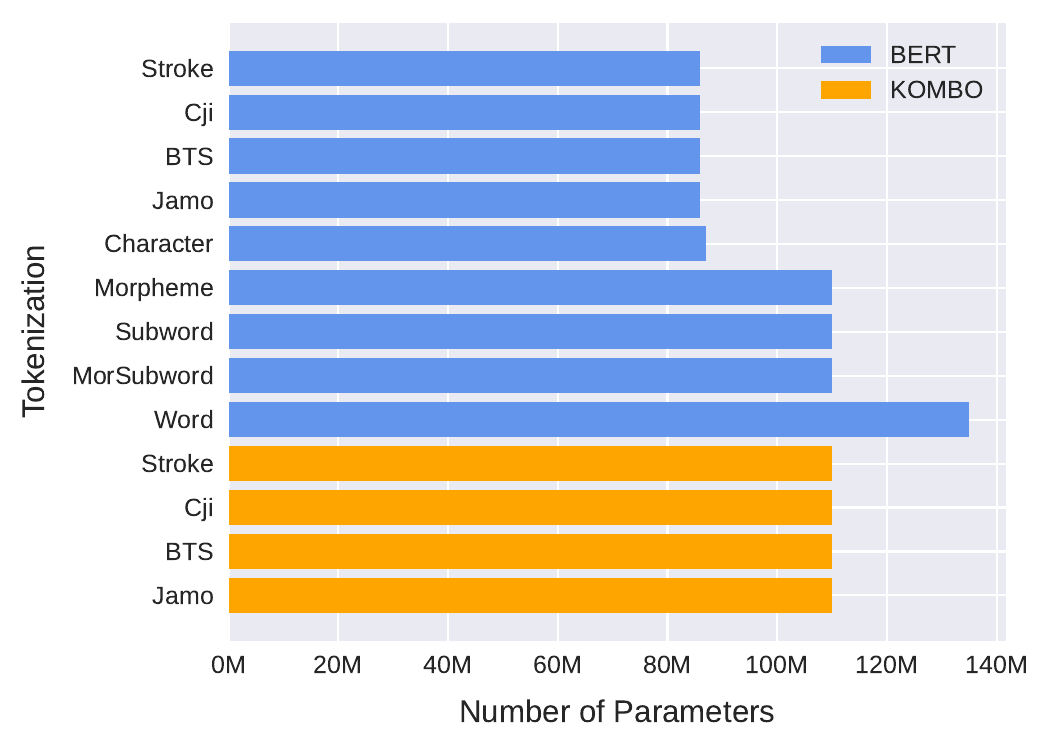}}
    \subfigure[]{\includegraphics[width=0.45\textwidth]{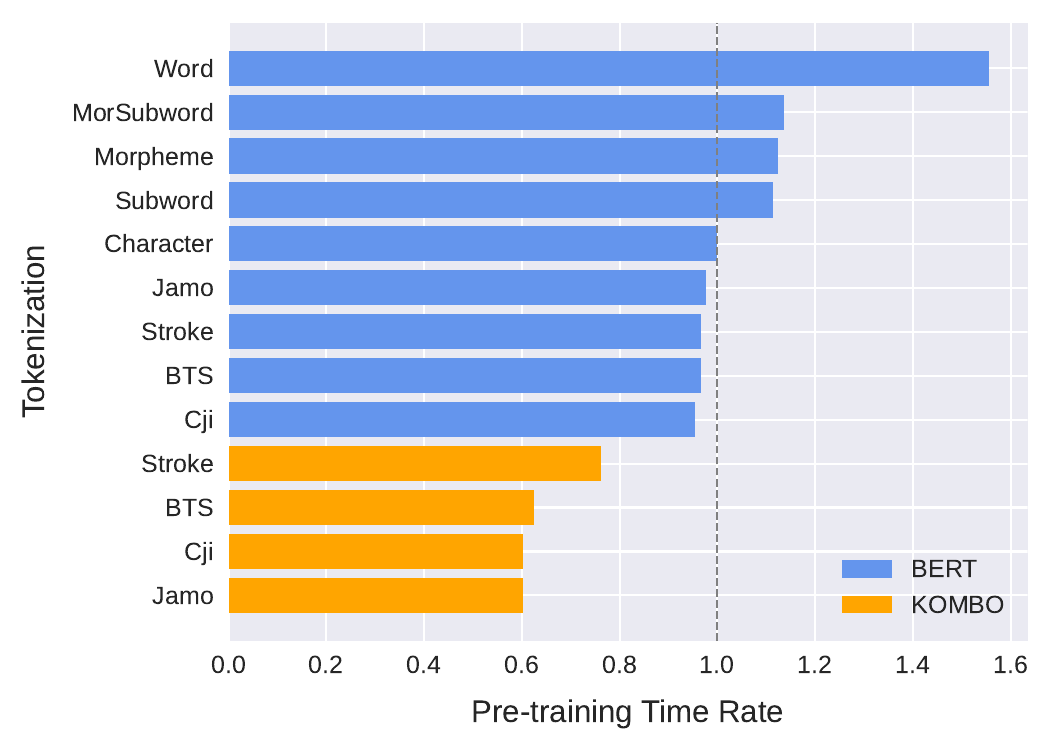}}
    \caption{Comparison of (a) number of parameters and (b) training time of model according to tokenization method for each model. To facilitate clear quantitative comparison of training times, we express the training time in rate.}
    \label{figure_5}
\end{figure}

\paragraph{Training Settings}~ Each model is pre-trained for 1M steps with a batch size of 128 using the RTX 3090 GPU. We set the AdamW optimizer \cite{loshchilov2018decoupled} with a learning rate of 5e-05 warm-up over the first 10K steps. We use a sequence length of 128. To match the size of our models with the state-of-the-art Korean pre-trained language model, MorSubword, we adjust the number of local transformer blocks in contextualization layers. The number of parameters in both models is 110M, but the training time of \raisebox{-0.1\height}{\includegraphics[height=0.9em]{assets/KOMBO.pdf}}$_\text{Jamo}$ is almost half of vanilla MorSubword model. We compare the number of parameters and the training time rate in Figure~\ref{figure_5}.

\section{Details of Fine-tuning}
\label{sec:appendixD}
\subsection{Standard Korean Datasets}
\label{sec:appendixD.1}
\paragraph{Datasets}~ KorQuAD 1.0 \cite{lim2019korquad1} is the Korean version of the well-known SQuAD 1.0 dataset \cite{rajpurkar-etal-2016-squad}, designed for the machine reading comprehension task. KorQuAD 1.0 dataset consists of 10,645 passages and 66,181 question-answer pairs (60,407 for the training set and 5,774 for the development set). Similar to SQuAD 1.0 dataset, the evaluation involves the model predicting the position of the answer within the given passage.

KorNLI \cite{ham-etal-2020-kornli} is a dataset consisting of 942,854 training sets and 7,500 evaluation sets for natural language inference, sourced from SNLI dataset \cite{bowman-etal-2015-large}, MNLI dataset \cite{williams-etal-2018-broad}, and XNLI dataset \cite{conneau-etal-2018-xnli}. The labels of the dataset are entailment, contradiction, and neutral.

KorSTS \cite{ham-etal-2020-kornli} is designed to evaluate the semantic similarity between two sentences. It is a Korean adaptation of STS-B dataset \cite{cer-etal-2017-semeval}. KorSTS dataset comprises 5,749 training examples and 2,879 evaluation examples. Each sample is labeled with a similarity score scaled from 0 to 5, indicating the similarity between the two sentences.

NSMC \cite{Park:2016} is a NAVER movie review dataset used for performing a sentiment analysis of each Korean sentence. NSMC dataset consists of 150,000 training samples and 50,000 test samples, with each sentence labeled as negative or positive.

PAWS-X \cite{yang-etal-2019-paws} is a dataset for the paraphrase identification task. PAWS-X dataset has six language tasks, and we only evaluate models on the Korean subset. It consists of 53,338 sentence pairs (49,410 for the training set, 1,965 for the development set, and 1,972 for the test set), and each label has two possible values, different meanings, or paraphrases.

\paragraph{Data Preprocessing}~ To focus solely on evaluating Korean language processing capabilities, we convert English words into \textit{Hangeul} based on the International Phonetic Alphabet (IPA) symbols~\footnote{To convert graphemes to phonemes, we leverage the g2pK library. Refer to \href{https://github.com/Kyubyong/g2pK}{https://github.com/Kyubyong/g2pK}} (e.g., `bus' → `버스') and remove special characters.

\paragraph{Hyperparameters}~ For each of the datasets, we use the following hyperparameters: KorQuAD 1.0 uses 5 epochs and a batch size of 16 . KorNLI uses 3 epochs and 16 batch size. KorSTS uses 5 epochs and a batch size of 64. NSMC uses 3 epochs and a batch size of 64. PAWS-X uses 5 epochs and a batch size of 64. We select the best learning rate (among 1e-04, 4e-05, and 5e-05) on the Dev set and warmup over the first 10\% steps of the total. We use a dropout probability of 0.1. We choose the maximum sequence length (among 128, 256, 512, 1024, and 2048) depending on the type of token.
% fine-tune our proposed models on the Korean NLU tasks with the hyperparameters, for which we selected the best learning settings: epochs = \{3, 5\}, batch size = \{16, 64\}, learning rate = \{1e-04, 4e-05, 5e-05\}, max sequence length = \{128, 256, 512, 1024, 2048\}, dropout = 0.1, and warmup rate= 0.1.

\subsection{Noisy Korean Datasets}
\label{sec:appendixD.2}
\begin{figure*}[t]
    \centering
    \subfigure[PAWS-X]{\includegraphics[width=0.32\textwidth]{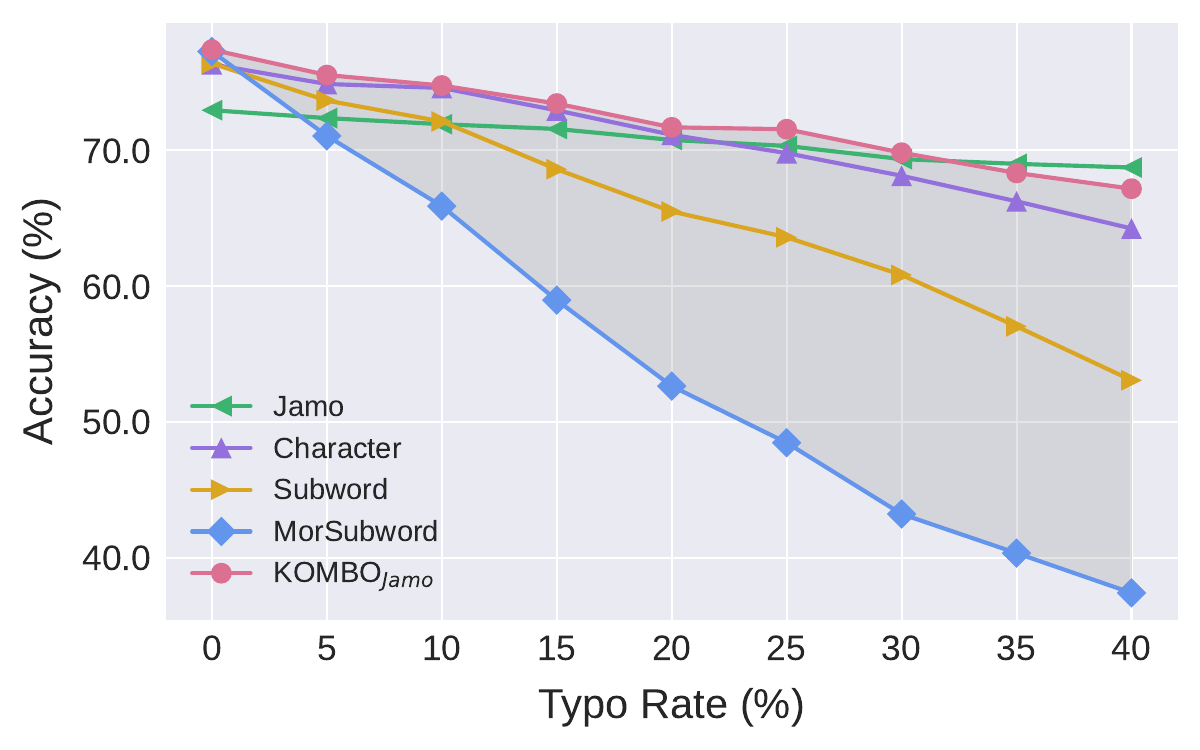}}
    \subfigure[NSMC]{\includegraphics[width=0.32\textwidth]{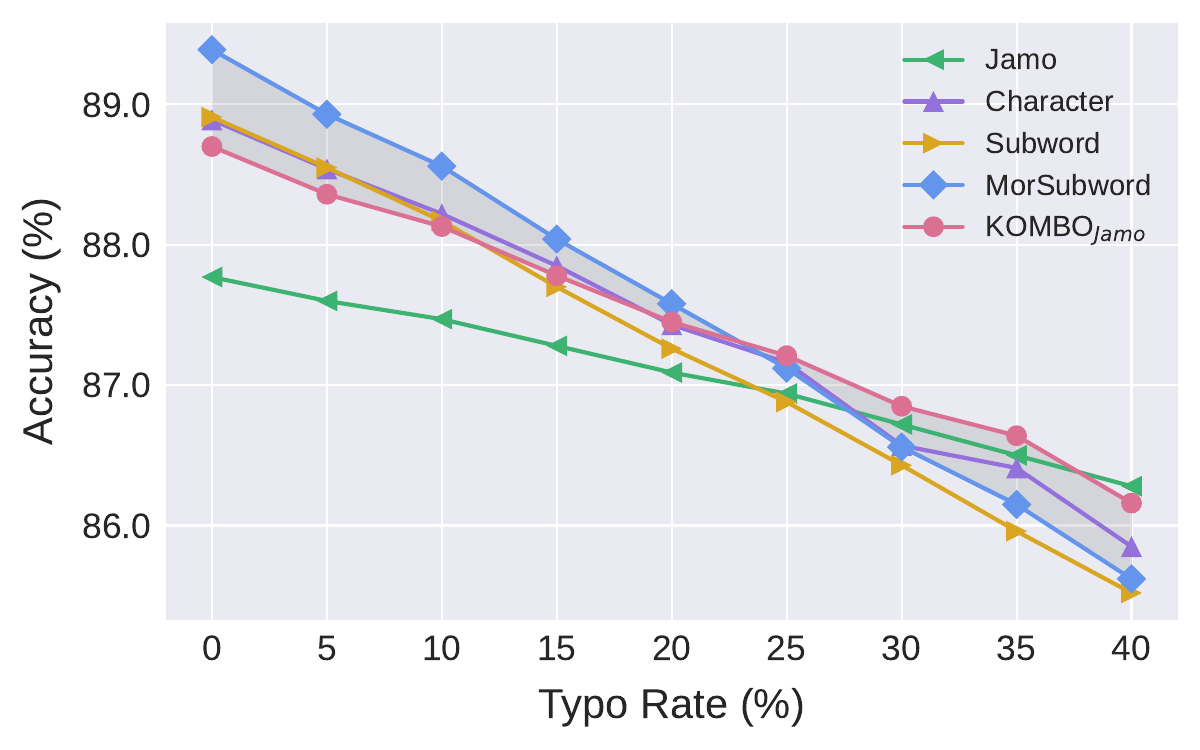}}
    \subfigure[PAWS-X]{\includegraphics[width=0.32\textwidth]{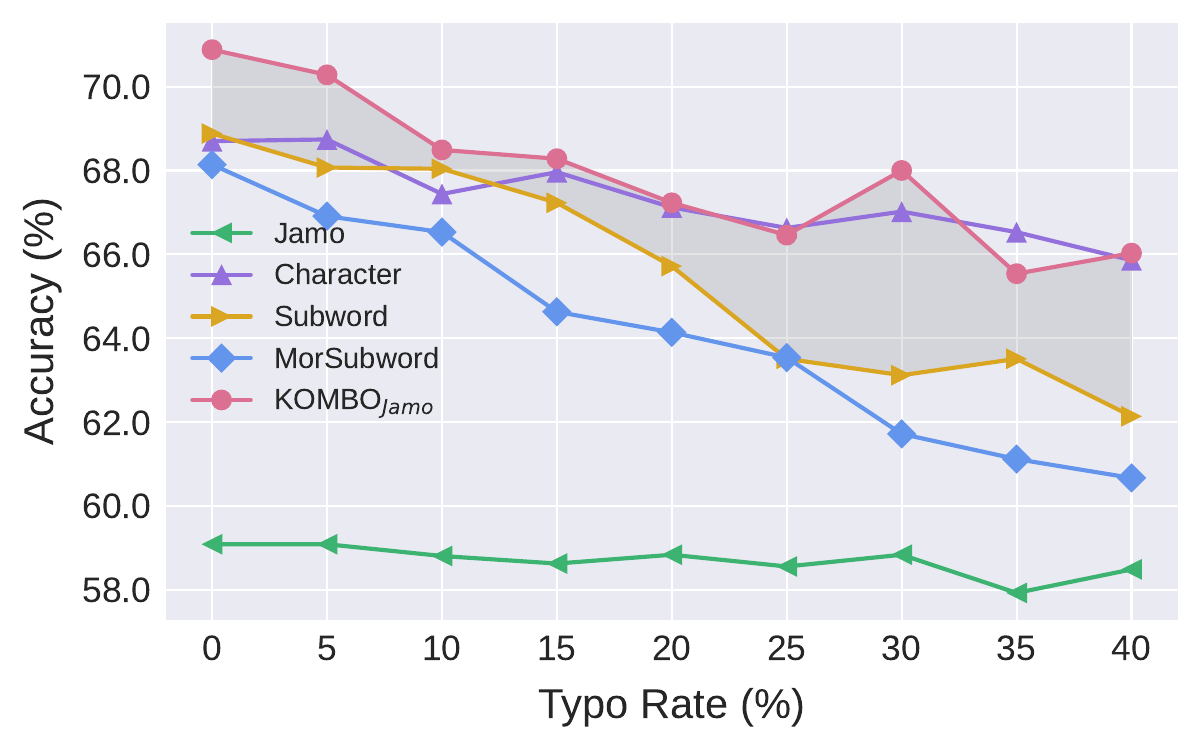}}
    \caption{Visualization of the evaluation results on KorSTS, NSMC, and PAWS-X datasets with increasing the typo rate. We fill the gap between our proposed method and the state-of-the-art baseline in gray.}
    \label{figure_6}
\end{figure*}

We conduct additional typo experiments for the KorSTS, NSMC, and PAWS-X datasets. In Figure~\ref{figure_6}, we illustrate the result of random typo experiments. In Table~\ref{table_7}, we present the results of each of the four distinct typo generation methods.

\subsection{Korean Offensive Language Dataset}
\label{sec:appendixD.3}
\paragraph{Datasets}~BEEP! \cite{moon-etal-2020-beep} is a Korean corpus annotated for toxic speech detection, which consists of 10K manually human-annotated Korean corpus collected from a Korean entertainment news aggregation platform. The labels include Hate, Offensive, and None, categorizing the aggressiveness of the given sentences. To distinguish whether the text contains offensive language or not, we reformulate the labels into `Hate or Offensive' and `None' for binary classification.

K-MHaS \cite{lee-etal-2022-k} is a Korean multi-label hate speech detection dataset that consists of 109K utterances from Korean news comments. K-MHaS requires the classification of fine-grained discrimination labels among Politics, Origin, Physical, Age, Gender, Religion, Race, and Profanity based on the input sentences.

KOLD \cite{jeong-etal-2022-kold} is a dataset for detecting the Korean offensive language. KOLD comprises 40,429 comments, annotated hierarchically with the type and the target. KOLD has three tasks: Level A (Offensive language detection), Level B (Target types categorization), and Level C (Target group Identification). In this paper, we implement the most difficult task, Level C in KOLD, labeling into 22 target groups: LGBTQ+, Men, Women, Asian, Black, Chinese, Indian, Korean-Chinese, Southeast Asian, White, Conservative, Progressive, Buddhism, Catholic, Christian, Islam, Agism, Disabled, Diseased, Feminist, Physical Appearance, and Socio-economic Status. 

For all offensive language datasets, we convert English words into \textit{Hangeul} based on the International Phonetic Alphabet (IPA) symbols.

\paragraph{Hyperparameters}~
We train the models on BEEP! for 10 epochs with 32 batch size. In the case of the K-MHaS dataset, we train the models for 4 epochs with 32 batch size. For the experiments on KOLD, we train the models for 5 epochs with a batch size of 32 using a learning rate of 5e-05.
We choose the max sequence length (among 128, 256, 512) depending on the type of tokens.

\section{Robustness to Character Conjugation}
\label{sec:appendixE}
\begin{figure*}[h]
    \centering
    \includegraphics[width=1.\textwidth]{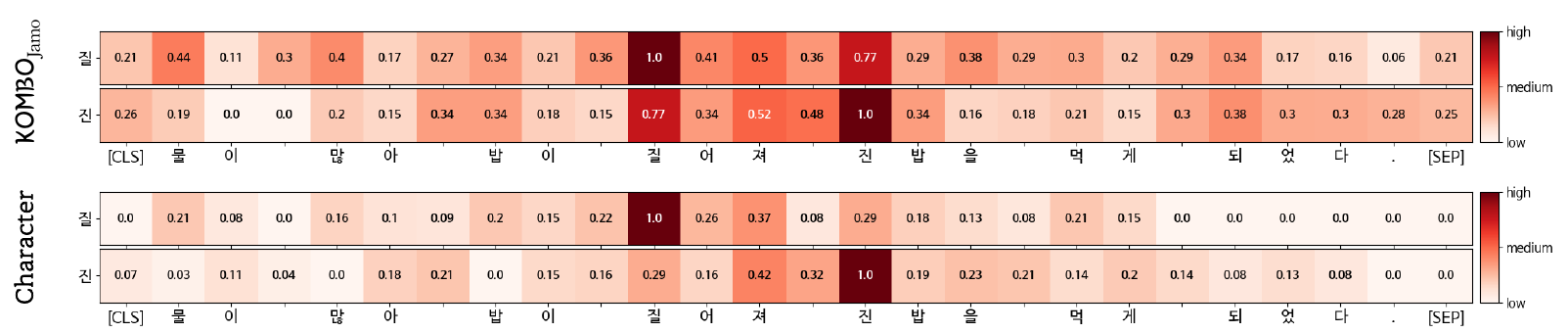}
    \caption{Visualization of the character representations.
    Given the target sentence as "물이 많아 밥이 \textbf{질}어져 \textbf{진}밥을 먹게 되었다. A lot of water makes the rice \textbf{mushy}, so I have to eat \textbf{mushy} rice.)", the histogram represents the cosine similarities between the embeddings of the character `질' and `진' extracted from the target sentence and all characters in the target sentence.}
    \label{figure_7}
\end{figure*}

\begin{figure*}[h]
    \centering
    \includegraphics[width=1.\textwidth]{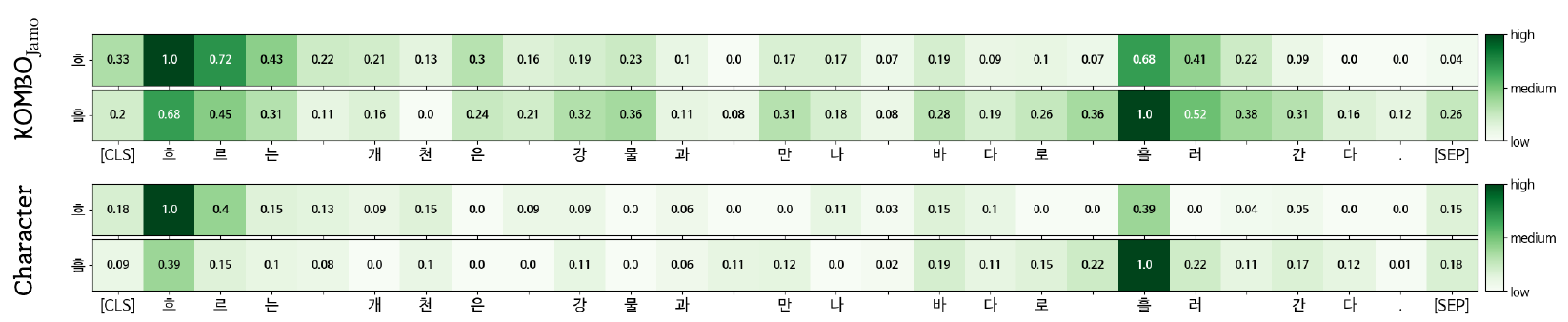}
    \caption{Visualization of the character representations.
    Given the target sentence as "\textbf{흐}르는 개천은 강물과 만나 바다로 \textbf{흘}러 간다. (A \textbf{flow}ing stream joins the river and \textbf{flow}s into the sea)", the histogram represents the cosine similarities between the embeddings of the character `흐' and `흘' extracted from the target sentence and all characters in the target sentence.}
    \label{figure_8}
\end{figure*}

\begin{figure*}[h]
    \centering
    \includegraphics[width=1.\textwidth]{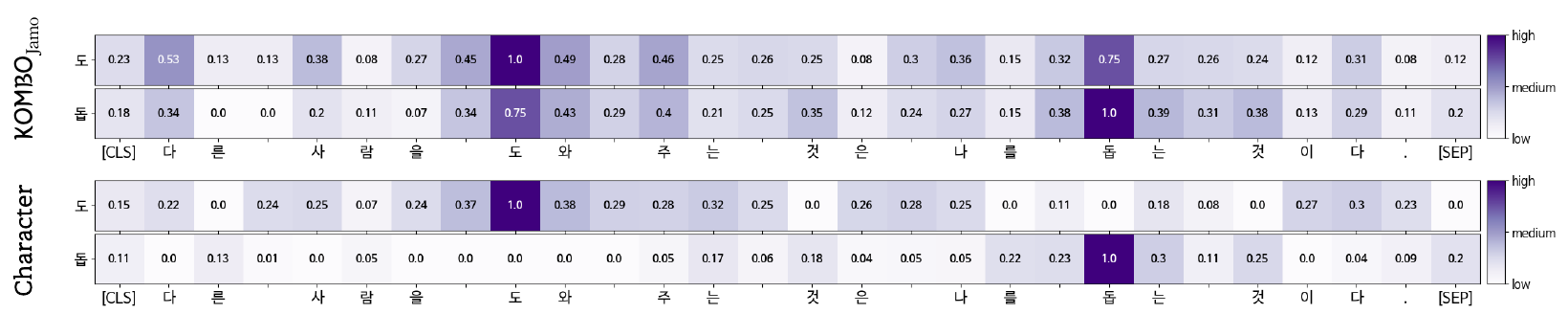}
    \caption{Visualization of the character representations.
    Given the target sentence as "다른 사람을 \textbf{도}와 주는 것은 나를 \textbf{돕}는 것이다. (\textbf{Help}ing others is to \textbf{help} myself.)", the histogram represents the cosine similarities between the embeddings of the character `도' and `돕' extracted from the target sentence and all characters in the target sentence.}
    \label{figure_9}
\end{figure*}

\begin{figure*}[h]
    \centering
    \includegraphics[width=1.\textwidth]{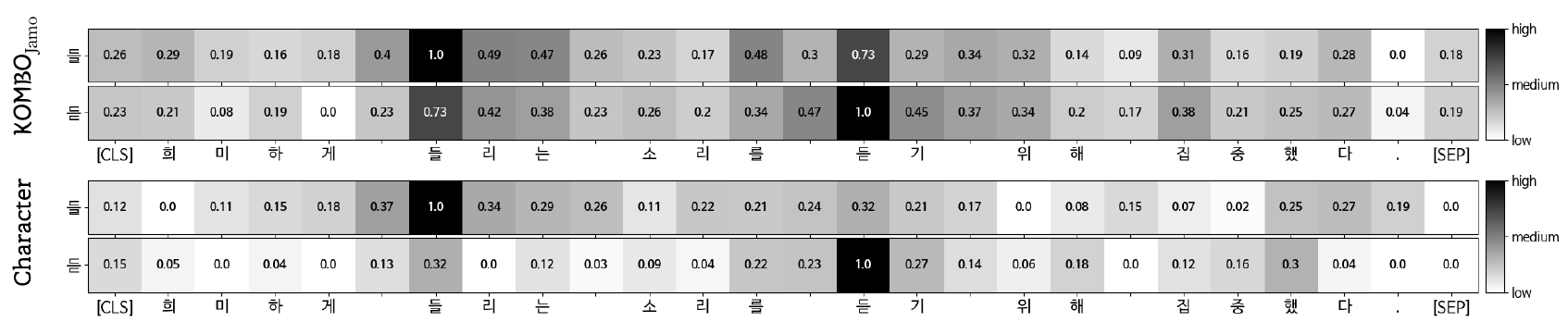}
    \caption{Visualization of the character representations.
    Given the target sentence as "희미하게 \textbf{들}리는 소리를 \textbf{듣}기 위해 집중했다. (I have to concentrate on \textbf{hear}ing the sound, which is unclear to \textbf{hear}.)", the histogram represents the cosine similarities between the embeddings of the character `들' and `듣' extracted from the target sentence and all characters in the target sentence. }
    \label{figure_10}
\end{figure*}

We illustrate more examples about Korean character conjugations in Figure~\ref{figure_7}, \ref{figure_8}, \ref{figure_9}, and \ref{figure_10}.

\begin{table*}[t]
% \resizebox{\textwidth}{!}{%
\centering
\begin{tabular}{clccccccccccc}
\toprule
\multicolumn{11}{c}{\textbf{KorSTS}}                       \\ 
\toprule
\multirow{2.5}{*}{Model} & \multirow{2.5}{*}{Tokenization} & \multirow{2.5}{*}{Clean} & \multicolumn{2}{c}{Insertion} & \multicolumn{2}{c}{Transposition} & \multicolumn{2}{c}{Substitution} & \multicolumn{2}{c}{Deletion} \\ 
 \cmidrule(lr){4-5} \cmidrule(lr){6-7} \cmidrule(lr){8-9} \cmidrule(lr){10-11}
 & & & +20\% & +40\% & +20\% & +40\% & +20\% & +40\% & +20\% & +40\% \\ 
\toprule
\multirow{4}{*}{$\text{BERT}$} & Jamo & 72.96 & \underline{71.67} & \textbf{69.61} & 47.68 & 29.77 & 72.15 & \textbf{69.62} & 47.31 & 30.84 \\ 
 & Character & 76.34 & 70.93 & 66.33 & 49.86 & \underline{32.30} & \underline{72.34} & \underline{68.95} & 49.63 & \underline{33.81} \\
 & Subword & 76.41 & 64.26 & 57.00 & 50.52 & 31.46 & 68.43 & 62.64 & 49.94 & 32.22 \\
 & MorSubword & \underline{77.29} & 66.09 & 57.26 & \textbf{57.21} & \textbf{34.89} & 68.04 & 57.42 & \textbf{57.09} & \textbf{36.29} \\
\midrule
\raisebox{-0.1\height}{\includegraphics[height=0.9em]{assets/KOMBO.pdf}} & Jamo & \textbf{77.42} & \textbf{73.13} & \underline{69.19} & \underline{52.34} & 31.64 & \textbf{73.21} & 68.17 & \underline{52.47} & 32.42 \\

\toprule
\multicolumn{11}{c}{\textbf{NSMC}} \\ \toprule
\multirow{2.5}{*}{Model} & \multirow{2.5}{*}{Tokenization} & \multirow{2.5}{*}{Clean} & \multicolumn{2}{c}{Insertion} & \multicolumn{2}{c}{Transposition} & \multicolumn{2}{c}{Substitution} & \multicolumn{2}{c}{Deletion} \\ 
 \cmidrule(lr){4-5} \cmidrule(lr){6-7} \cmidrule(lr){8-9} \cmidrule(lr){10-11}
 & & & +20\% & +40\% & +20\% & +40\% & +20\% & +40\% & +20\% & +40\% \\ 
\toprule
\multirow{4}{*}{$\text{BERT}$} & Jamo & 87.78 & 87.22 & \textbf{86.58} & 84.04 & 79.17 & 87.22 & \underline{86.68} & 84.08 & 79.45 \\ 
 & Character & 88.89 & \textbf{87.69} & 86.20 & 84.72 & 79.44 & 87.62 & 86.24 & \underline{84.95} & \underline{79.89} \\
 & Subword & \underline{88.91} & 87.41 & 85.86 & \underline{84.84} & \underline{79.56} & 87.31 & 85.57 & 84.90 & 79.73 \\
 & MorSubword & \textbf{89.40} & \underline{87.57} & 85.72 & \textbf{85.30} & \textbf{79.94} & \underline{87.66} & 85.63 & \textbf{85.29} & \textbf{79.93} \\
\midrule
\raisebox{-0.1\height}{\includegraphics[height=0.9em]{assets/KOMBO.pdf}} & Jamo & 88.70 & \underline{87.57} & \underline{86.33} & 84.49 & 79.15 & \textbf{87.80} & \textbf{86.70} & 84.64 & 79.44 \\

\toprule
\multicolumn{11}{c}{\textbf{PAWS-X}}
\\
\toprule
\multirow{2.5}{*}{Model} & \multirow{2.5}{*}{Tokenization} & \multirow{2.5}{*}{Clean} & \multicolumn{2}{c}{Insertion} & \multicolumn{2}{c}{Transposition} & \multicolumn{2}{c}{Substitution} & \multicolumn{2}{c}{Deletion} \\ 
 \cmidrule(lr){4-5} \cmidrule(lr){6-7} \cmidrule(lr){8-9} \cmidrule(lr){10-11}
 & & & +20\% & +40\% & +20\% & +40\% & +20\% & +40\% & +20\% & +40\% \\ 
\toprule
\multirow{4}{*}{$\text{BERT}$} & Jamo & 59.09 & 58.32 & 57.65 & 56.95 & \underline{56.07} & 58.18 & 58.84 & 58.07 & 56.77 \\ 
 & Character & 68.70 & \underline{66.91} & \underline{64.91} & 60.28 & \textbf{56.98} & \underline{66.88} & \underline{65.33} & 61.33 & 57.16 \\
 & Subword & \underline{68.88} & 64.63 & 61.23 & \underline{60.63} & 55.40 & 66.60 & 62.60 & 61.37 & 56.84 \\
 & MorSubword & 68.14 & 62.88 & 60.56 & 60.25 & 55.40 & 63.51 & 60.67 & \underline{61.86} & \textbf{57.65} \\
\midrule
\raisebox{-0.1\height}{\includegraphics[height=0.9em]{assets/KOMBO.pdf}} & Jamo & \textbf{70.88} & \textbf{69.30} & \textbf{67.44} & \textbf{63.02} & 55.26 & \textbf{69.09} & \textbf{67.44} & \textbf{63.51} & \underline{57.51} \\

\toprule
\end{tabular}
\caption{Evaluation results on the typo dataset. We measure the sensitivity to typo rate using four different generation methods.The best and second-best results are highlighted in \textbf{boldface} and \underline{underline}, respectively.}
\label{table_7}
\end{table*}

\clearpage

\end{CJK}
\end{document}